\documentclass{article}
\PassOptionsToPackage{table,xcdraw}{xcolor}
\usepackage{microtype}
\usepackage{graphicx}
\usepackage{subcaption}
\usepackage{xcolor}
\usepackage{booktabs} % for professional tables
\usepackage[T1]{fontenc}
\usepackage{hyperref}
\usepackage{multirow}
\usepackage{enumitem}

\definecolor{R1}{HTML}{56C596}         % 绿色
\definecolor{R1-Searcher}{HTML}{4DA5D9} % 浅蓝
\definecolor{Search-R1}{HTML}{3C88C6}   % 蓝色
\definecolor{Graph-R1}{HTML}{6A0DAD} 

\usepackage[accepted]{icml2026}

\usepackage{amsmath}
\usepackage{amssymb}
\usepackage{mathtools}
\usepackage{amsthm}
\usepackage{algorithm}
\usepackage{algpseudocode}

\usepackage[capitalize,noabbrev]{cleveref}

\theoremstyle{plain}

\theoremstyle{definition}

\theoremstyle{remark}

\usepackage[textsize=tiny]{todonotes}

\icmltitlerunning{Graph-R1: Towards Agentic GraphRAG Framework via End-to-end Reinforcement Learning}

\begin{document}

\twocolumn[
  \icmltitle{Graph-R1: Towards Agentic GraphRAG Framework via End-to-end Reinforcement Learning}



  \icmlsetsymbol{equal}{*}

\begin{icmlauthorlist}
\icmlauthor{Haoran Luo}{bupt,ntu}
\icmlauthor{Haihong E}{bupt}
\icmlauthor{Guanting Chen}{bupt}
\icmlauthor{Qika Lin}{nus}
\icmlauthor{Yikai Guo}{hk2y}
\icmlauthor{Fangzhi Xu}{xjtu}
\icmlauthor{Zemin Kuang}{anzhen}
\icmlauthor{Meina Song}{bupt}
\icmlauthor{Xiaobao Wu}{sjtu}
\icmlauthor{Yifan Zhu}{bupt}
\icmlauthor{Luu Anh Tuan}{ntu,vin}
\end{icmlauthorlist}

\icmlaffiliation{bupt}{Beijing University of Posts and Telecommunications}
\icmlaffiliation{ntu}{Nanyang Technological University}
\icmlaffiliation{nus}{National University of Singapore}
\icmlaffiliation{hk2y}{Beijing Institute of Computer Technology and Application}
\icmlaffiliation{xjtu}{Xi'an Jiaotong University}
\icmlaffiliation{anzhen}{Beijing Anzhen Hospital, Capital Medical University}
\icmlaffiliation{sjtu}{Shanghai Jiao Tong University}
\icmlaffiliation{vin}{VinUniversity}

% \icmlfirstauthor{Haoran Luo}{haoran.luo@ieee.org}

\icmlcorrespondingauthor{Haihong E}{ehaihong@bupt.edu.cn}

  \icmlkeywords{Machine Learning, ICML}

  \vskip 0.3in
]



\printAffiliationsAndNotice{}  % no special notice (required even if empty)

\begin{abstract}
Retrieval-Augmented Generation (RAG) mitigates hallucination in LLMs by incorporating external knowledge, but relies on chunk-based retrieval that lacks structural semantics. GraphRAG methods improve RAG by modeling knowledge as entity-relation graphs, but still face challenges in high construction cost, fixed one-time retrieval, and reliance on long-context reasoning and prompt design. To address these challenges, we propose \texttt{\textbf{Graph-R1}}, the first agentic GraphRAG framework via end-to-end reinforcement learning (RL). It introduces lightweight knowledge hypergraph construction, models retrieval as a multi-turn agent-environment interaction, and optimizes the agent process via an end-to-end reward mechanism. Experiments on standard RAG datasets show that Graph-R1 outperforms traditional GraphRAG and RL-enhanced RAG methods in reasoning accuracy, retrieval efficiency, and generation quality. Our software and data are publicly available at \url{https://github.com/LHRLAB/Graph-R1}.
\end{abstract}

\section{Introduction}

Large Language Models (LLMs)~\citep{LLMsurvey} have achieved widespread success in NLP tasks. However, when applied to knowledge-intensive or proprietary knowledge-dependent applications, they still suffer from the hallucination problem~\citep{Hsurvey}, generating inaccurate content. To improve credibility and factual consistency, Retrieval-Augmented Generation (RAG)~\citep{RAG} introduces external knowledge sources as references, alleviating the knowledge bottleneck of pure language modeling. Nevertheless, existing RAG methods mostly rely on chunk-based text blocks~\citep{RAGsurvey}, which makes it difficult to capture complex knowledge structures among entities. To address this, GraphRAG methods~\citep{GraphRAG,LightRAG,HyperGraphRAG} represent knowledge as entity-relation graphs, enhancing retrieval efficiency.

\begin{figure}[t]
\centering
\includegraphics[width=8.2cm]{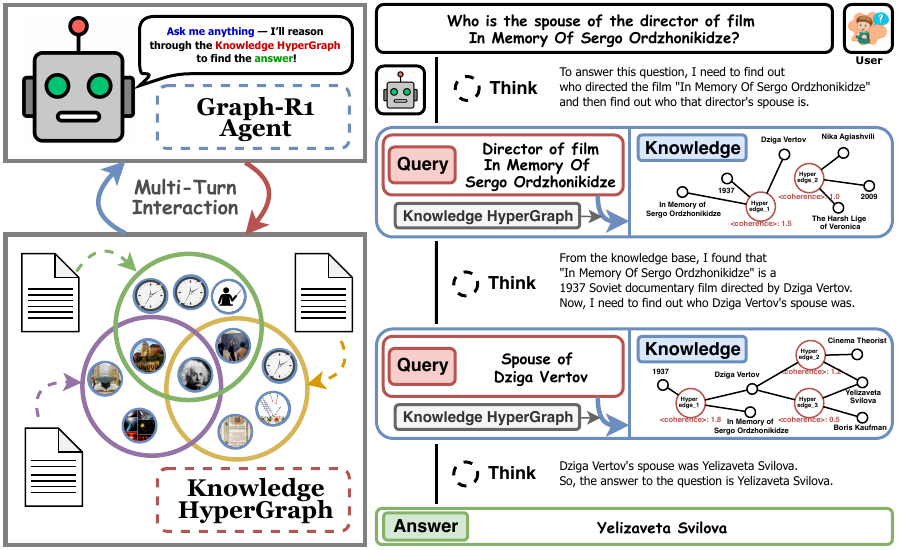}
\caption{
An illustration of Graph-R1. Graph-R1 formulates GraphRAG as a multi-turn agentic retrieval process over a knowledge hypergraph, optimized via end-to-end reinforcement learning.
}
\label{F1}
% \vspace{-0.8mm}
\end{figure}

Generally, GraphRAG methods consist of three processes: \textit{knowledge graph construction}, \textit{graph retrieval}, and \textit{answer generation}. First, knowledge graphs are typically constructed by LLMs to extract entities and relations from text, forming a graph structure~\citep{Esurvey}. Second, the retrieval process queries relevant subgraphs or paths through subgraph retrieval or path pruning strategies~\citep{PathRAG,HippoRAG2}. Finally, the generation process prompts LLMs to generate answers based on the retrieved graph-based knowledge~\citep{GraphRAG-Bench}.

\begin{figure*}[t]
\centering
\includegraphics[width=0.95\linewidth]{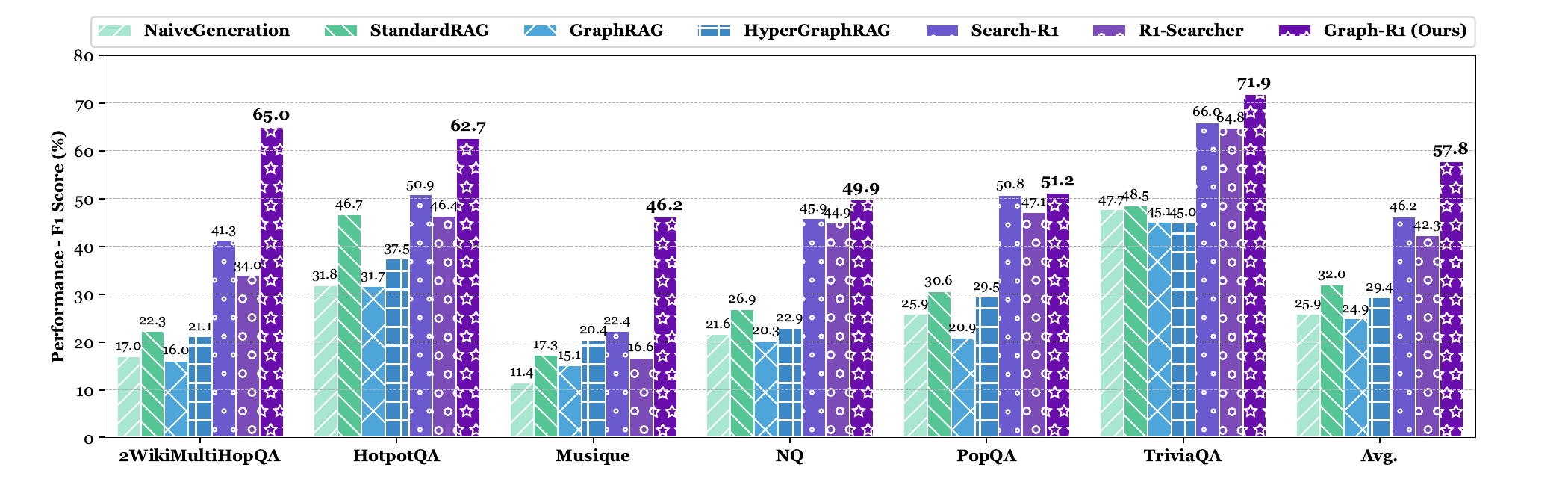}
\caption{Comparison of F1 scores across RAG benchmarks. Using a graph as the knowledge environment enables RL to achieve a higher performance ceiling compared to chunk-based knowledge. Graph-structured knowledge boosts accuracy by richer representation.}
\label{F2}
\end{figure*}

However, current GraphRAG methods still face three key challenges:
\textbf{(i) High cost and semantic loss in \textit{knowledge construction} process.} 
Compared to standard RAG, GraphRAG methods convert natural language knowledge into graph structures using LLMs, which results in high cost and often causes semantic loss relative to the original content~\citep{Text2NKG,HyperGraphRAG}.
% {\color{red}
% \begingroup
% \textbf{(ii) Fixed retrieval process with only one-time interaction in \textit{graph retrieval}.} 
% Although most existing GraphRAG methods design various retrieval strategies to improve efficiency, they still rely on a \emph{single-pass or fixed retrieval pipeline} to gather sufficient knowledge~\citep{PathRAG}, which limits their ability to correct missing or noisy evidence in complex queries.
% \endgroup}
\textbf{(ii) Fixed retrieval process with only one-time interaction in \textit{graph retrieval} process.} 
Although existing GraphRAG methods design various retrieval strategies to improve efficiency, most of them aim to gather sufficient knowledge in a single fixed retrieval~\citep{PathRAG}, which limits performance in complex queries.
\textbf{(iii) Dependence on large LLMs for long-context analysis and prompt quality in \textit{answer generation} process.} 
Generation based on retrieved graph-structured knowledge often requires strong long-context reasoning ability, making the output quality highly dependent on the LLM's parameter size and prompt design~\citep{LightRAG}.

To address these challenges, we propose \texttt{\textbf{Graph-R1}}, as illustrated in Figure~\ref{F1}, the first agentic GraphRAG framework enhanced by end-to-end reinforcement learning (RL), inspired by DeepSeek-R1~\citep{DeepSeek-R1}. First, we propose a lightweight knowledge hypergraph construction method to establish a standard agent environment for the query action space. Moreover, we model the retrieval process as a multi-turn agentic interaction process, enabling LLMs to repeatedly perform the reasoning loop of “think-retrieve-rethink-generate” within the knowledge hypergraph environment. Furthermore, we design an end-to-end reward mechanism that integrates generation quality, retrieval relevance, and structural reliability of graph paths into a unified optimization objective. Using RL, the agent learns a generalizable graph reasoning strategy and achieves tighter alignment between structured knowledge and language.

We perform experiments on various standard RAG datasets~\citep{FlashRAG}. Experimental results demonstrate that Graph-R1 outperforms traditional GraphRAG methods and RAG combined with RL methods~\citep{Search-R1,R1-Searcher} in reasoning accuracy, retrieval efficiency, and generation quality. As shown in Figure~\ref{F2}, the end-to-end RL strategy guides the agent through multiple turns of interaction and goal-driven exploration in the graph, effectively bridging the gap between knowledge representation and language generation. This work lays a foundation for building the next generation of knowledge-driven and strategy-optimized agent-based generation systems.

\section{Related Work}

\textbf{RAG and GraphRAG. }
Retrieval-Augmented Generation (RAG)~\citep{RAG} improves LLM factuality by retrieving external knowledge, but struggles with data silos and limited structural reasoning. GraphRAG~\citep{GraphRAG} mitigates these issues by incorporating graph-structured knowledge, inspiring enterprise systems~\citep{MedGraphRAG,KAG,PIKE-RAG} and efficient variants like LightRAG~\citep{LightRAG}. Recent works further enhance representation power by diverse graph structures~\citep{HyperGraphRAG,Hyper-RAG,HyperRec,CausalRAG,NodeRAG,LinearRAG}, while retrieval optimization explores path-based exploration and pruning techniques~\citep{PathRAG,HippoRAG2,HopRAG,PropRAG,E2GraphRAG,GFM-RAG,G-reasoner,HybGRAG,Youtu-GraphRAG,GraphSearch}. In this work, we propose Graph-R1, the first agentic GraphRAG framework that learns a multi-turn retrieval policy via end-to-end RL.

\textbf{Reinforcement Learning for LLMs. }
Reinforcement learning (RL) is increasingly adopted to enhance LLM reasoning~\citep{R1survey,KBQA-o1}, as demonstrated by OpenAI’s o1/o3/o4~\citep{GPT-o1}. DeepSeek-R1~\citep{DeepSeek-R1} achieves comparable capabilities and introduces the Group Relative Policy Optimization (GRPO)~\citep{GRPO} for scalable end-to-end training, which has been extended to various downstream tasks \citep{SQL-R1,Psy-R1,AutoBM}. RL-enhanced agents have also shown strong performance in multi-turn interaction with different environments~\citep{KBQA-o1,GiGPO,Search-R1,R1-Searcher,Prompt-R1}, highlighting RL’s potential in advancing agentic GraphRAG framework~\citep{RAGR1survey}.

\begin{figure*}[t]
\centering
\includegraphics[width=0.89\linewidth]{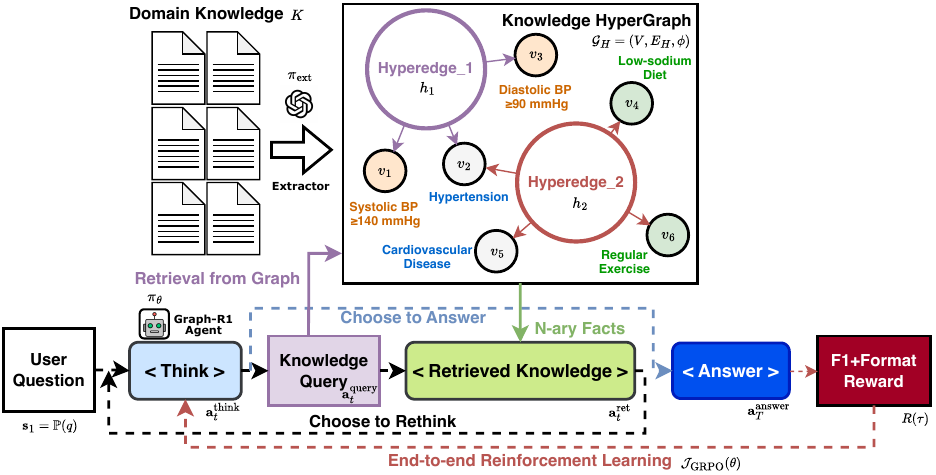}
\caption{Overview of the Graph-R1 framework: an RL-enhanced reasoning trajectory over knowledge hypergraph, where the agent iteratively decides to think, query from graph environment, retrieve n-ary facts as candidate knowledge, and answer the user question.}
\label{F3}
\end{figure*}

\section{Preliminaries}

We formalize the GraphRAG pipeline into three stages:

\textbf{(a) Knowledge Graph Construction.}
Given a knowledge collection $K = \{d_1, d_2, \dots, d_N\}$, the goal is to extract facts $f_d$ from each semantic unit $d \in K$ into a unified graph $\mathcal{G}_K$:
\begin{equation}
\small
\mathcal{G}_K \sim \sum_{d \in K} \pi_{\text{ext}}(f_d \mid d),
\end{equation}
where $\pi_{\text{ext}}$ denotes an LLM-based extractor that parses each $d$ into a set of relation-entity pairs $f_d = \{(r_i, \mathcal{V}_{r_i})\}$, with $r_i$ as the relation and $\mathcal{V}_{r_i} = \{v_1, \dots, v_n\}$ the entities.

\textbf{(b) Graph Retrieval.}
Graph retrieval is formulated as a two-step process over $\mathcal{G}_K$: (1) retrieving candidate reasoning paths and (2) pruning irrelevant ones. Conditioned on a query $q$, the model first retrieves a candidate set $\mathcal{X}_q = \{x_1, \dots, x_m\}$ and then selects a relevant subset $\mathcal{Z}_q \subseteq \mathcal{X}_q$.
The overall objective is to maximize the joint likelihood:
\begin{equation}
\small
\begin{aligned}
&\max_{\theta} \; \mathbb{E}_{\mathcal{Z}_q \sim P(\mathcal{Z}_q \mid q, \mathcal{G}_K)} \\
&\left[
\prod_{t=1}^{T_x} P_\theta(x_t \mid x_{<t}, q, \mathcal{G}_K) \cdot
\prod_{t=1}^{T_z} P_\theta(z_t \mid z_{<t}, \mathcal{X}_q, q)
\right],
\end{aligned}
\end{equation}
where \(T_x\) \& \(T_z\) are numbers of retrieved \& selected paths.

\textbf{(c) Answer Generation.}
Given a query $q$ and selected paths $\mathcal{Z}_q$, answer generation produces a natural language answer $y$ grounded in graph-based evidence, formulated as:
\begin{equation}
\small
P(y \mid q, \mathcal{G}_K) = \sum_{\mathcal{Z}_q \subseteq \mathcal{X}_q} P(y \mid q, \mathcal{Z}_q) \cdot P(\mathcal{Z}_q \mid q, \mathcal{G}_K),
\end{equation}
where $P(y \mid q, \mathcal{Z}_q)$ is generation likelihood and $P(\mathcal{Z}_q \mid q, \mathcal{G}_K)$ is retrieval-pruning distribution.

\begin{table*}[t]
\caption{\label{T1}
Template for Graph-R1. \textcolor{red}{question} will be replaced with the specific user query. Note that the knowledge retrieved is placed within \textcolor{teal}{\texttt{\textbf{<knowledge>}}}...\textcolor{teal}{\texttt{\textbf{</knowledge>}}} after \textcolor{orange}{\texttt{\textbf{<query>}}}...\textcolor{orange}{\texttt{\textbf{</query>}}}.}
\centering
% \vskip -0.2in
\small
\begin{tabular}{p{16.7cm}}
\toprule
You are a helpful assistant. Answer the given question. 
You can query from knowledge base provided to you to answer the question. 
You can query knowledge as many times as you want. 
You must first conduct reasoning inside \textcolor{blue}{\texttt{\textbf{<think>}}}...\textcolor{blue}{\texttt{\textbf{</think>}}}. 
If you need to query knowledge, you can set a query statement between 
\textcolor{orange}{\texttt{\textbf{<query>}}}...\textcolor{orange}{\texttt{\textbf{</query>}}} 
to query from knowledge base after \textcolor{blue}{\texttt{\textbf{<think>}}}...\textcolor{blue}{\texttt{\textbf{</think>}}}.  
When you have the final answer, you can output the answer inside 
\textcolor{purple}{\texttt{\textbf{<answer>}}}...\textcolor{purple}{\texttt{\textbf{</answer>}}}. 
Question: \textcolor{red}{question}. Assistant: \\
\bottomrule
\end{tabular}
\end{table*}

\section{Methodology: Graph-R1}
In this section, as illustrated in Figure~\ref{F3}, we introduce Graph-R1, including agent initialization, multi-turn graph interaction, and outcome-directed end-to-end RL optimization.

\subsection{Agent Initialization}
\label{Section4.1}
Graph-R1 adopts an LLM-driven agent, initialized with a knowledge hypergraph environment $\mathcal{G}_H$, the action space $\mathcal{A}$, the state space $\mathcal{S}$, and the answer target $y_q$, given $q$.

\textbf{Graph Environment $\mathcal{G}_H$.}
We propose a lightweight method for constructing a knowledge hypergraph $\mathcal{G}_H$ from given domain knowledge $K = \{d_1, d_2, \dots, d_N\}$. For each chunk unit $d \in K$, an LLM-based extractor $\pi_{\text{ext}}$ identifies $m$ n-ary relational facts, where each comprises a semantic segment $h_i$ and a set of participating entities $\mathcal{V}_{h_i} = \{v_1, \dots, v_n\}$. A shared encoder $\phi(\cdot)$ is then used to generate semantic embeddings for both entities \& relations:
\begin{equation}
\small
\label{E4}
\begin{aligned}
&\mathcal{G}_H = (V, E_H, \phi), \ \  \text{where}\ \  \pi_{\text{ext}}(d) \rightarrow \{(h_i, \mathcal{V}_{h_i})\}_{i=1}^{m}, \\
&\text{and}\ \ \phi(v) = \mathrm{Enc}(v),\; \phi(h_i) = \mathrm{Enc}(h_i),
\end{aligned}
\end{equation}
where each $h_i$ defines a hyperedge $h_i \in E_H$ connecting its associated entities $\mathcal{V}_{h_i}$ as $v \in V$. The resulting hypergraph $\mathcal{G}_H$ encodes n-ary relations with rich semantic grounding.

\textbf{The Agent Action Space $\mathcal{A}$.}
In Graph-R1, each agent action $\mathbf{a}_t \in \mathcal{A}$ comprises four sub-actions: \textit{Thinking} $\mathbf{a}_t^{\text{think}}$, which decides whether to continue or terminate reasoning; \textit{Query Generation} $\mathbf{a}_t^{\text{query}}$, which formulates a retrieval query; \textit{Graph Retrieval} $\mathbf{a}_t^{\text{ret}}$, which extracts relevant knowledge from the hypergraph; and \textit{Answering} $\mathbf{a}_t^{\text{ans}}$, which produces a final response if reasoning ends.

The agent action $\mathbf{a}_t$ has two compositional forms, and the joint action log-likelihood $\log \pi(\mathbf{a}_t \mid \mathbf{s}_t)$ is defined as:
\begin{equation}
\small
\begin{aligned}
&\begin{cases}
\log \mathcal{G}_H(\mathbf{a}_t^{\text{ret}} \mid \mathbf{s}_t, \mathbf{a}_t^{\text{think}}, \mathbf{a}_t^{\text{query}}) + \log \pi(\mathbf{a}_t^{\text{query}} \mid \mathbf{s}_t, \mathbf{a}_t^{\text{think}}) + \\[-0.1ex]
\log \pi(\mathbf{a}_t^{\text{think}} \mid \mathbf{s}_t), & \!\!\!\!\!\!\!\!\!\!\!\!\!\!\!\!\!\!\!\!\!\!\!\!\!\!\!\!\!\!\!\!\!\!\!\!\!\!\!\!\!\text{if } \mathbf{a}_t^{\text{think}} \rightarrow
 \text{continue}, \\[0.5ex]
\log \pi(\mathbf{a}_t^{\text{ans}} \mid \mathbf{s}_t, \mathbf{a}_t^{\text{think}}) + \log \pi(\mathbf{a}_t^{\text{think}} \mid \mathbf{s}_t), & \!\!\!\!\!\!\!\!\!\!\!\!\!\!\!\!\!\!\!\!\!\!\!\!\!\!\!\!\!\!\!\!\!\!\!\!\!\!\!\!\!\text{if } \mathbf{a}_t^{\text{think}} \rightarrow
 \text{terminate},
\end{cases}
\end{aligned}
\end{equation}
where, at each step, the agent first performs \textit{Thinking}, and then chooses between continuing reasoning (\textit{Query Generation} and \textit{Graph Retrieval}) or terminating via \textit{Answering}.

\textbf{The Agent State Space $\mathcal{S}$ and Target $y_q$.}
At each step $t$, the state $\mathbf{s}_t \in \mathcal{S}$ is defined as $\mathbf{s}_t = (\mathbf{s}_1, \mathbf{a}_1, \dots, \mathbf{a}_{t-1})$, with $\mathbf{s}_1$ initialized from the input query $q$. Once a termination action $\mathbf{a}_T$ is issued, the agent reaches final state $\mathbf{s}_T$, where $T$ is the total number of reasoning steps, and an answer $y_q \sim \mathbf{a}_T^{\text{ans}}$ is produced to address $q$.

\textbf{Proposition 1.} \textit{Graph-structured knowledge boosts agent accuracy by richer representation.}\vspace{-3mm}
\begin{proof}
We provide quantitative experimental results in Section~\ref{5.2} and qualitative proofs in Appendix~\ref{proof1}.
\end{proof}

\begin{figure*}[t]
\footnotesize
\centering

\begin{subfigure}{0.29\textwidth}
  \includegraphics[width=\linewidth]{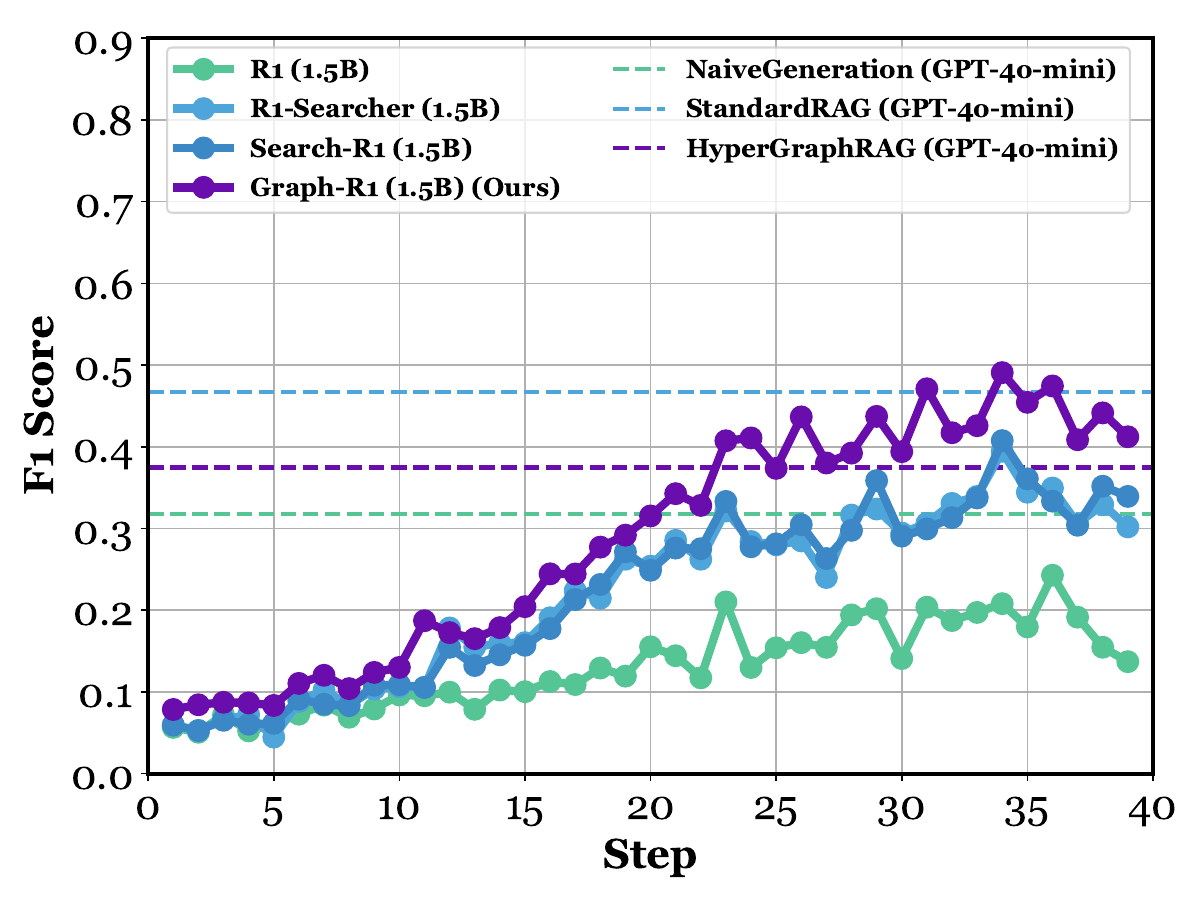}
  \vspace{-4.5mm}
  \caption{Qwen2.5-1.5B-Instruct}
  \label{F4a}
\end{subfigure}
\hspace{1mm}
\begin{subfigure}{0.29\textwidth}
  \includegraphics[width=\linewidth]{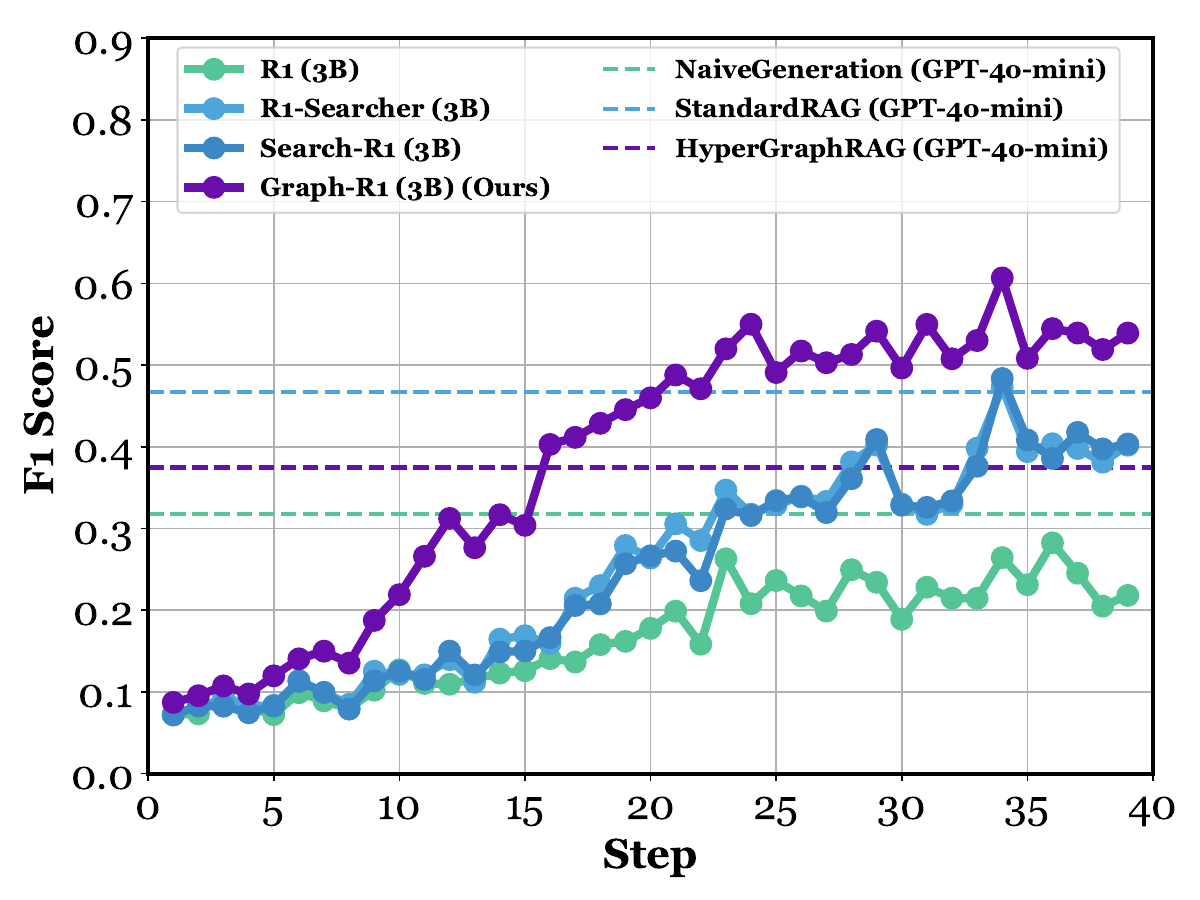}
  \vspace{-4.5mm}
  \caption{Qwen2.5-3B-Instruct}
  \label{F4b}
\end{subfigure}
\hspace{1mm}
\begin{subfigure}{0.29\textwidth}
  \includegraphics[width=\linewidth]{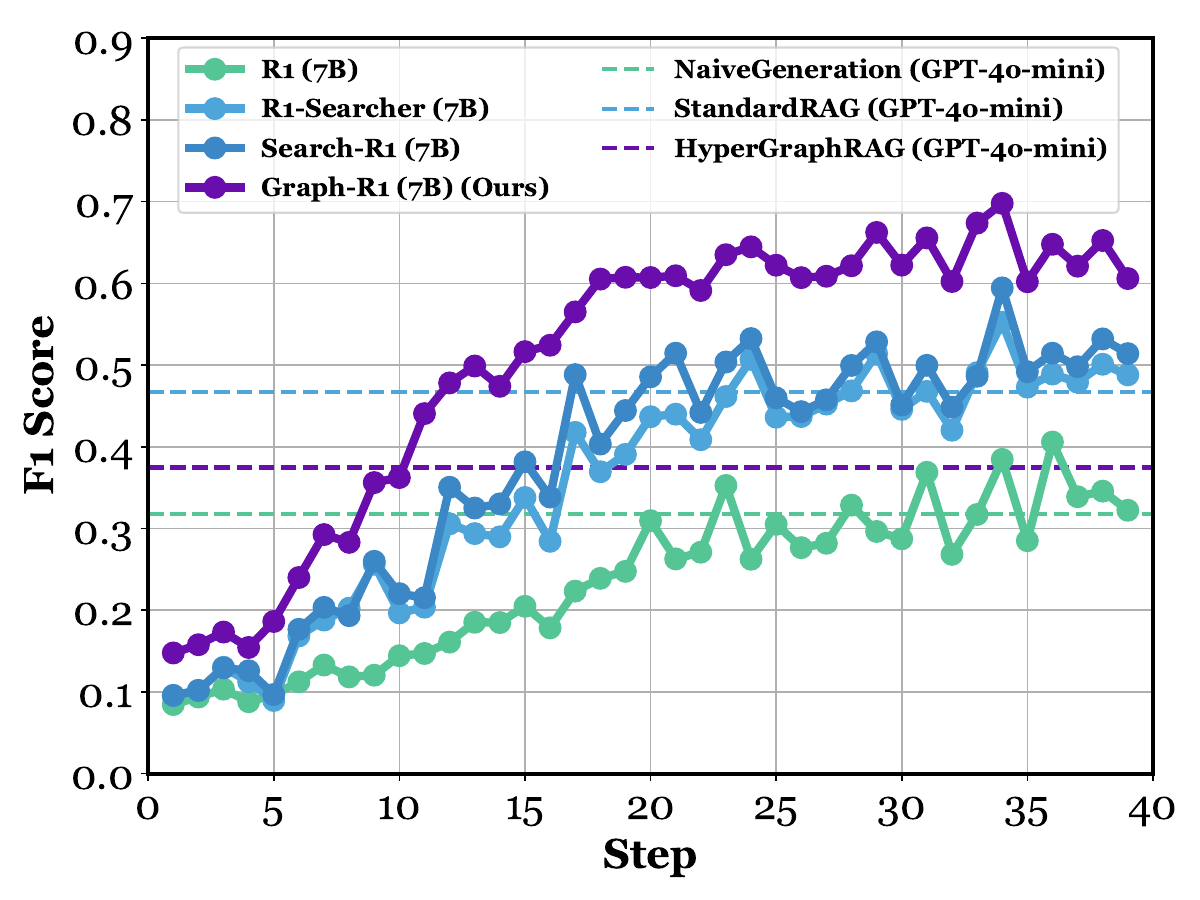}
  \vspace{-4.5mm}
  \caption{Qwen2.5-7B-Instruct}
  \label{F4c}
\end{subfigure}

\vspace{-1mm}
\caption{Step-wise F1 score based on Qwen2.5 (1.5B, 3B, 7B), where Graph-R1 outperforms baselines and GPT-4o-mini variants.}
\label{F4}
\end{figure*}

\subsection{Reasoning via Multi-turn Graph Interaction}
We model reasoning as a multi-turn interaction between an agent $\pi_\theta$ and a hypergraph $\mathcal{G}_H$. We first define the step-wise policy $\pi_\theta(\cdot \mid \mathbf{s}_t)$ prompted by Table~\ref{T1}, then describe how to retrieve knowledge $\mathcal{G}_H(\mathbf{a}_t^{\text{ret}} \mid \cdot, \mathbf{a}_t^{\text{query}})$ based on $\mathbf{a}_t^{\text{query}}$, and finally present the objective to optimize $P(y_q \mid \cdot)$.

\textbf{Modeling the Step-wise Reasoning Policy.}  
At each reasoning step $t$, the LLM governs the agent’s behavior by generating a structured output consisting of: (i) a thinking reflection $\mathbf{a}_t^{\text{think}}$ that summarizes the current state and highlights potential knowledge gaps;
(ii) a composition indicator $\alpha_t \in \mathcal{A}_{\text{type}} = \{\texttt{(query, retrieve)}, \texttt{(answer)}\}$ that determines the sub-action structure; and  
(iii) a content output $\mathbf{a}_t^{\text{out}} \in \mathcal{A}_{\text{content}}$, representing either a retrieval query or a final answer.
We model this decision-making process as a hierarchical policy conditioned on the agent state $\mathbf{s}_t \in \mathcal{S}$. The policy is factorized as:
\begin{equation}
\label{E6}
\small
\begin{aligned}
&\pi_\theta(\mathbf{a}_t^{\text{think}}, \alpha_t, \mathbf{a}_t^{\text{out}} \mid \mathbf{s}_t) =\\
&\pi_\theta(\mathbf{a}_t^{\text{out}} \mid \alpha_t, \mathbf{a}_t^{\text{think}}, \mathbf{s}_t) \cdot
\pi_\theta(\alpha_t \mid \mathbf{a}_t^{\text{think}}, \mathbf{s}_t) \cdot
\pi_\theta(\mathbf{a}_t^{\text{think}} \mid \mathbf{s}_t),
\end{aligned}
\end{equation}
where $\pi_\theta$ denotes the LLM-parameterized policy, which encourages three aligned behaviors: generating reflections $\mathbf{a}_t^{\text{think}}$ that assess knowledge sufficiency, selecting $\alpha_t$ to balance exploration and termination, and producing $\mathbf{a}_t^{\text{out}}$ that advances retrieval $\mathbf{a}_t^{\text{query}}$ or yields a direct answer $\mathbf{a}_t^{\text{ans}}$.

\textbf{Knowledge Interaction via Hypergraph Retrieval.}  
Given a query $\mathbf{a}_t^{\text{query}}$ generated by the reasoning LLM, we retrieve relevant knowledge $\mathbf{a}_t^{\text{ret}}$ from the hypergraph $\mathcal{G}_H = (V, E_H)$ through a dual-path interaction process: entity-based retrieval and direct hyperedge retrieval. The resulting n-ary relational facts are then aggregated via rank-based fusion to support downstream reasoning.

\textit{(i) Entity-based Hyperedge Retrieval.}  
We first identify a set of top-ranked entities based on their similarity to the extracted entities $V_{\mathbf{a}_t^{\text{query}}}$, and collect hyperedges that connect to any retrieved entity:
\begin{equation}
\small
\begin{aligned}
&\mathcal{R}_V(\mathbf{a}_t^{\text{query}}) = \operatorname*{argmax}^{k_V}_{v \in V} \left(\text{sim}(\phi(V_{\mathbf{a}_t^{\text{query}}}),\phi(v))\right), \\
&\text{and}\ \ \mathcal{F}_V^*= \bigcup_{v_i \in \mathcal{R}_V} \{ (e_H, V_{e_H}) \mid v_i \in V_{e_H}, e_H \in E_H \},
\end{aligned}
\end{equation}
where $\phi(V_{\mathbf{a}_t^{\text{query}}})$ is the aggregated embedding of entities extracted from $\mathbf{a}_t^{\text{query}}$, $\phi(v)$ is the entity embedding, $k_V$ is the number of retrieved entities, and $V_{e_H}$ denotes the entity set of hyperedge $e_H$.

\textit{(ii) Direct Hyperedge Retrieval.}  
In parallel, we directly retrieve hyperedges based on query-hyperedge similarity, and collect their associated relational facts:
\begin{equation}
\small
\begin{aligned}
&\mathcal{R}_H(\mathbf{a}_t^{\text{query}}) = \operatorname*{argmax}^{k_H}_{e_H \in E_H} \left(\text{sim}(\phi(\mathbf{a}_t^{\text{query}}), \phi(e_H))\right),\\
&\text{and}\ \ \mathcal{F}_H^* = \bigcup_{e_i \in \mathcal{R}_H} \{ (e_i, V_{e_i}) \mid V_{e_i} \subseteq V \},
\end{aligned}
\end{equation}
where $\phi(\mathbf{a}_t^{\text{query}})$ is the query embedding, $\phi(e_H)$ is the hyperedge embedding, $k_H$ is the number of retrieved hyperedges, and $V_{e_i}$ denotes the entity set of hyperedge $e_i$.

\textit{(iii) Fusion via Reciprocal Rank Aggregation.}  
To produce the final knowledge set, we merge results from both retrieval paths using reciprocal rank aggregation over hyperedges:
\begin{equation}
\small
\mathbf{a}_t^{\text{ret}} = \mathcal{F}_{\mathbf{a}_t^{\text{query}}}^* = \text{Top-}k\left( \mathcal{F}_V^* \cup \mathcal{F}_H^*, \;\text{Score}(f) = \frac{1}{r_V} + \frac{1}{r_H} \right),
\end{equation}
where $r_V$ and $r_H$ are the ranks of n-ary relational fact $f$ in $\mathcal{F}_V^*$ and $\mathcal{F}_H^*$ respectively (set to $\infty$ if absent), and $k$ is the number of retrieved facts $\mathbf{a}_t^{\text{ret}}$ returned to the agent.

\textbf{Optimization Objective for Agent Trajectories.}  
The agent aims to learn a reasoning trajectory $\tau \in \mathcal{T}_q$ that yields a faithful and contextually grounded answer $y_q$. Each trajectory $\tau = ((\mathbf{s}_1, \mathbf{a}_1), (\mathbf{s}_2, \mathbf{a}_2), \dots, (\mathbf{s}_T, \mathbf{a}_T))$ comprises a sequence of actions executed over $\mathcal{G}_H$, defined as:
\begin{equation}
\small
\max_{\theta} \;\; \mathbb{E}_{\tau \sim \pi_\theta(\mathcal{T}_q \mid q; \mathcal{G}_H)} \left[ \log P(y_q \mid \tau) \right],
\end{equation}
where $P(y_q \mid \tau)$ denotes the likelihood of the correct answer $y_q \sim \mathbf{a}_t^{\text{ans}}$ under trajectory $\tau$, guiding $\pi_\theta$ toward answer-consistent reasoning.

\textbf{Proposition 2.} \textit{Multi-turn interaction with the graph environment improves retrieval efficiency.}\vspace{-3mm}
\begin{proof}
We provide quantitative experimental results in Section~\ref{5.5} and qualitative proofs in Appendix~\ref{proof2}.
\end{proof}

\subsection{Outcome-directed End-to-end RL Optimization}
To optimize the reasoning policy $\pi_\theta$ toward generating faithful and well-structured answers, we adopt an end-to-end reinforcement learning objective based on Group Relative Policy Optimization (GRPO)~\citep{GRPO} $\mathcal{J}_{\text{GRPO}}(\theta)$ and design an outcome-directed reward function $R(\tau)$. 

\begin{table*}[t]
\caption{\label{T2}
Main results with best in \textbf{bold}. \raisebox{-0.5mm}{\includegraphics[width=0.02\textwidth]{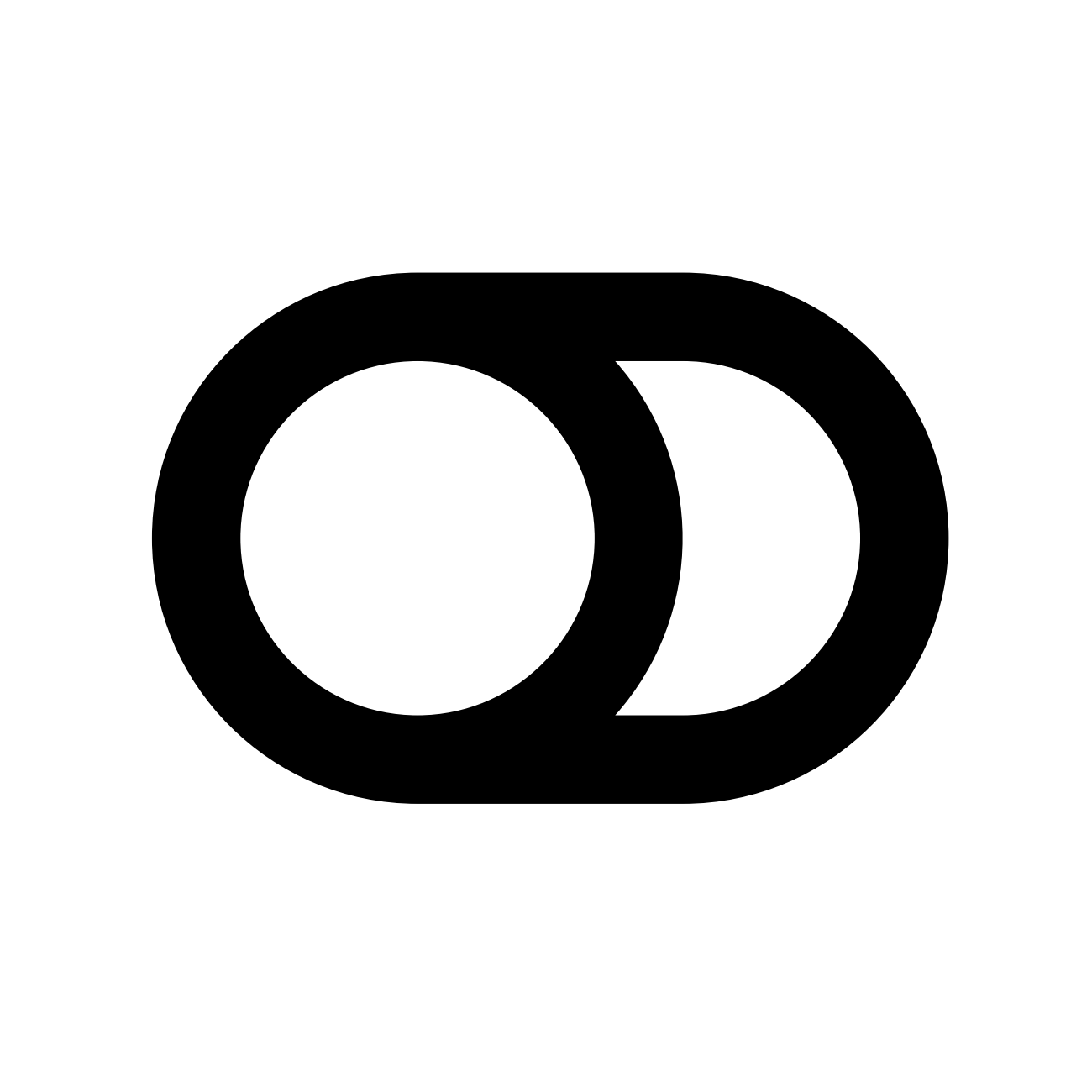}} means prompt engineering, \raisebox{-0.5mm}{\includegraphics[width=0.02\textwidth]{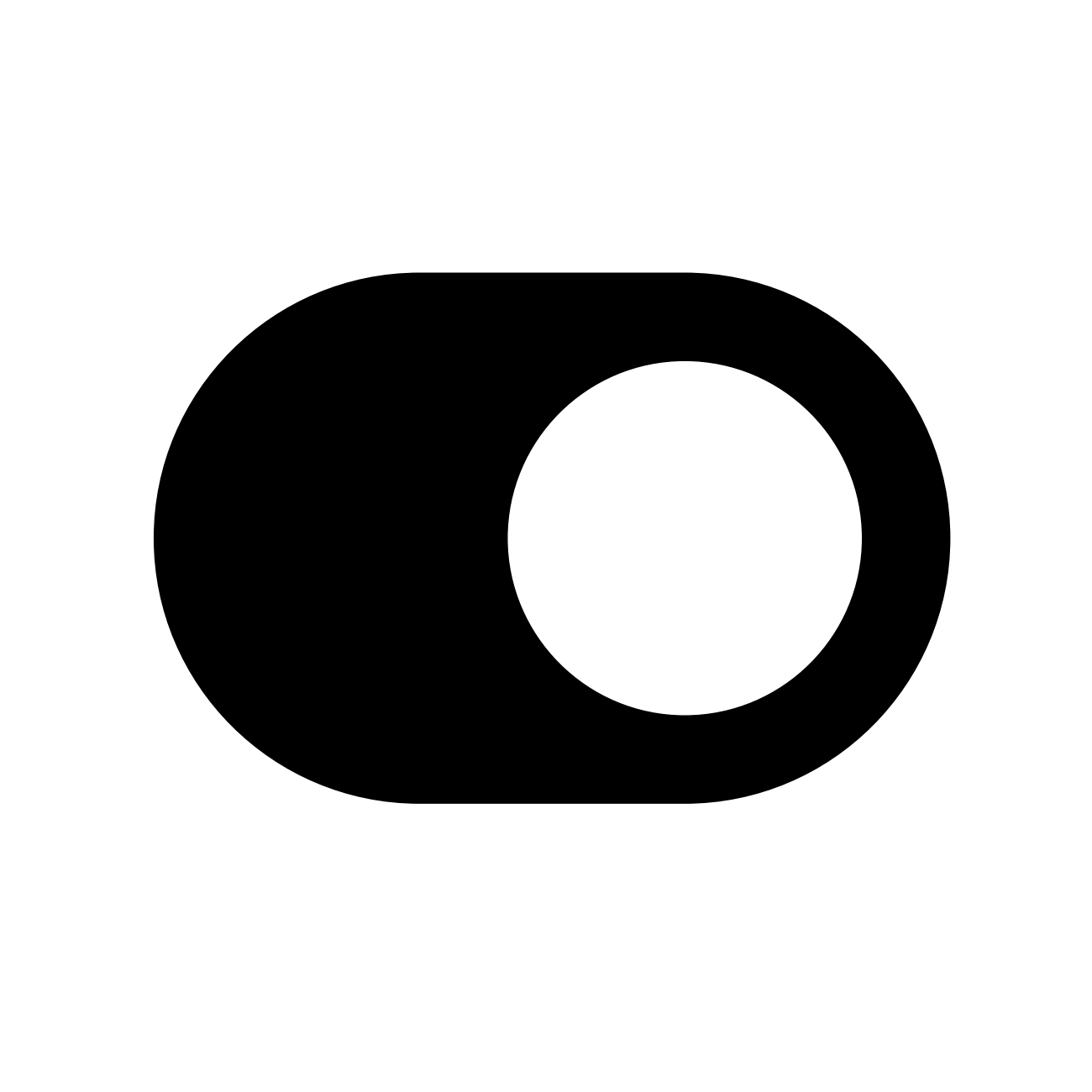}} means training, \raisebox{-0.5mm}{\includegraphics[width=0.02\textwidth]{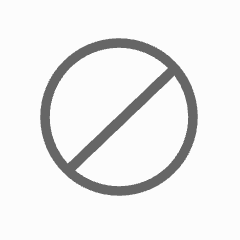}} means no knowledge interaction, \raisebox{-0.5mm}{\includegraphics[width=0.02\textwidth]{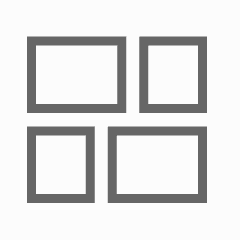}} means chunk-based knowledge, and \raisebox{-0.5mm}{\includegraphics[width=0.02\textwidth]{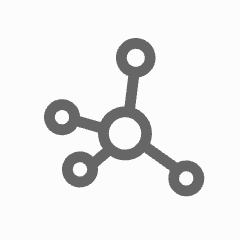}} means graph-based knowledge.}
\centering
\fontsize{7pt}{-10.5pt}\selectfont
\setlength{\tabcolsep}{1.7mm}{
\begin{tabular}{l cc  cc cc cc cc cc |cccc}
\toprule
\multirow{2.5}{*}{\textbf{Method}} & \multicolumn{2}{c}{\textbf{2Wiki.}} & \multicolumn{2}{c}{\textbf{HotpotQA}} & \multicolumn{2}{c}{\textbf{Musique}} & \multicolumn{2}{c}{\textbf{NQ}} & \multicolumn{2}{c}{\textbf{PopQA}} & \multicolumn{2}{c}{\textbf{TriviaQA}} & \multicolumn{4}{c}{\textbf{Avg.}} \\
\cmidrule(lr){2-3} \cmidrule(lr){4-5} \cmidrule(lr){6-7} \cmidrule(lr){8-9} \cmidrule(lr){10-11} \cmidrule(lr){12-13} \cmidrule(lr){14-17}
 & \textbf{F1} & \textbf{G-E} & \textbf{F1} & \textbf{G-E} & \textbf{F1} & \textbf{G-E} & \textbf{F1} & \textbf{G-E} & \textbf{F1} & \textbf{G-E} & \textbf{F1} & \textbf{G-E} & \textbf{EM} & \textbf{F1} & \textbf{R-S} & \textbf{G-E}\\
\midrule
\multicolumn{17}{c}{\textbf{\textit{GPT-4o-mini}}} \\
\raisebox{-0.22\height}{\includegraphics[width=0.02\textwidth]{close.png}}\raisebox{-0.22\height}{\includegraphics[width=0.02\textwidth]{none.png}} NaiveGeneration & 17.03 & 74.86 & 31.79 & 78.48 & 11.45 & 76.61 & 21.59 & 84.64 & 25.95 & \textbf{72.75} & 47.73 & 83.33 & 11.36 & 25.92 & - & 78.45 \\
\raisebox{-0.22\height}{\includegraphics[width=0.02\textwidth]{close.png}}\raisebox{-0.22\height}{\includegraphics[width=0.02\textwidth]{chunk.png}} StandardRAG & \textbf{22.31} & 73.02 & \textbf{46.70} & \textbf{81.88} & 17.31 & 74.93 & \textbf{26.85} & 84.55 & \textbf{30.58} & 69.42 & \textbf{48.55} & 84.63 & \textbf{18.10} & \textbf{32.05} & 52.68 & 78.07 \\
\raisebox{-0.22\height}{\includegraphics[width=0.02\textwidth]{close.png}}\raisebox{-0.22\height}{\includegraphics[width=0.02\textwidth]{graph.png}} GraphRAG & 16.02 & 72.81 & 31.67 & 77.37 & 15.14 & 74.43 & 20.31 & 82.36 & 20.92 & 65.88 & 45.13 & 82.76 & 12.50 & 24.87 & 32.48 & 75.94 \\
\raisebox{-0.22\height}{\includegraphics[width=0.02\textwidth]{close.png}}\raisebox{-0.22\height}{\includegraphics[width=0.02\textwidth]{graph.png}} LightRAG & 16.59 & 71.94 & 30.70 & 73.42 & 14.39 & 73.75 & 19.09 & 80.20 & 20.47 & 67.76 & 40.18 & 81.60 & 9.77 & 23.57 & 47.42 & 74.78 \\
\raisebox{-0.22\height}{\includegraphics[width=0.02\textwidth]{close.png}}\raisebox{-0.22\height}{\includegraphics[width=0.02\textwidth]{graph.png}} PathRAG & 12.42 & 67.19 & 23.12 & 71.81 & 11.49 & 69.94 & 20.01 & 81.99 & 15.65 & 60.58 & 37.44 & 80.94 & 7.03 & 20.02 & 46.71 & 72.08 \\
\raisebox{-0.22\height}{\includegraphics[width=0.02\textwidth]{close.png}}\raisebox{-0.22\height}{\includegraphics[width=0.02\textwidth]{graph.png}} HippoRAG2 & 16.27 & 68.78 & 31.78 & 76.43 & 12.37 & 73.05 & 24.56 & \textbf{84.65} & 21.10 & 63.31 & 46.86 & 83.55 & 13.80 & 25.49 & 36.41 & 74.96 \\
\raisebox{-0.22\height}{\includegraphics[width=0.02\textwidth]{close.png}}\raisebox{-0.22\height}{\includegraphics[width=0.02\textwidth]{graph.png}} HyperGraphRAG & 21.14 & \textbf{76.76} & 37.46 & 80.50 & \textbf{20.40} & \textbf{79.29} & 22.95 & 81.22 & 29.48 & 70.55 & 44.95 & \textbf{85.20} & 13.15 & 29.40 & \textbf{61.82} & \textbf{78.92} \\
\midrule
\multicolumn{17}{c}{\textbf{\textit{Qwen2.5-1.5B-Instruct}}} \\
\raisebox{-0.22\height}{\includegraphics[width=0.02\textwidth]{close.png}}\raisebox{-0.22\height}{\includegraphics[width=0.02\textwidth]{none.png}} NaiveGeneration & 7.78 & 49.13 & 4.27 & 45.77 & 2.35 & 46.63 & 6.03 & 46.74 & 10.06 & 42.67 & 8.10 & 52.92 & 1.17 & 6.43 & - & 47.31 \\
\raisebox{-0.22\height}{\includegraphics[width=0.02\textwidth]{close.png}}\raisebox{-0.22\height}{\includegraphics[width=0.02\textwidth]{chunk.png}} StandardRAG & 11.46 & 55.38 & 9.93 & 52.91 & 3.18 & 39.46 & 11.39 & 59.73 & 13.08 & 50.29 & 17.43 & 60.52 & 5.73 & 11.08 & 52.84 & 53.05 \\
\raisebox{-0.22\height}{\includegraphics[width=0.02\textwidth]{open.png}}\raisebox{-0.22\height}{\includegraphics[width=0.02\textwidth]{none.png}} SFT & 13.26 & 34.72 & 13.61 & 38.93 & 5.14 & 28.50 & 11.56 & 46.61 & 15.61 & 31.35 & 26.18 & 46.66 & 9.83 & 14.23 & - & 37.80 \\
\rowcolor{R1!15} \raisebox{-0.22\height}{\includegraphics[width=0.02\textwidth]{open.png}}\raisebox{-0.22\height}{\includegraphics[width=0.02\textwidth]{none.png}} R1 & 26.28 & 47.48 & 20.07 & 44.43 & 4.84 & 39.12 & 16.75 & 45.95 & 21.36 & 44.50 & 34.78 & 48.59 & 14.19 & 20.68 & - & 45.01 \\
\rowcolor{Search-R1!15} \raisebox{-0.22\height}{\includegraphics[width=0.02\textwidth]{open.png}}\raisebox{-0.22\height}{\includegraphics[width=0.02\textwidth]{chunk.png}} Search-R1 & 28.43 & 60.61 & 39.99 & 64.16 & 4.69 & 39.32 & 20.26 & 59.93 & 39.63 & 58.19 & 44.16 & 63.01 & 23.18 & 29.53 & 50.45 & 57.54 \\
\rowcolor{R1-Searcher!15} \raisebox{-0.22\height}{\includegraphics[width=0.02\textwidth]{open.png}}\raisebox{-0.22\height}{\includegraphics[width=0.02\textwidth]{chunk.png}} R1-Searcher & 28.01 & 58.81 & \textbf{41.50} & 61.54 & 6.26 & 38.31 & \textbf{36.86} & \textbf{60.79} & 38.37 & 56.02 & 42.57 & 61.24 & 23.70 & 32.26 & 50.68 & 56.12 \\
\rowcolor{Graph-R1!15} \raisebox{-0.22\height}{\includegraphics[width=0.02\textwidth]{open.png}}\raisebox{-0.22\height}{\includegraphics[width=0.02\textwidth]{graph.png}} Graph-R1 (ours) & \textbf{35.13} & \textbf{65.73} & 40.62 & \textbf{65.30} & \textbf{28.28} & \textbf{58.82} & 35.62 & 59.13 & \textbf{43.55} & \textbf{66.46} & \textbf{57.36} & \textbf{70.83} & \textbf{31.90} & \textbf{40.09} & \textbf{59.35} & \textbf{64.38}\\
\midrule
\multicolumn{17}{c}{\textbf{\textit{Qwen2.5-3B-Instruct}}} \\
\raisebox{-0.22\height}{\includegraphics[width=0.02\textwidth]{close.png}}\raisebox{-0.22\height}{\includegraphics[width=0.02\textwidth]{none.png}} NaiveGeneration & 7.59 & 55.00 & 11.16 & 53.75 & 3.67 & 54.00 & 8.90 & 57.18 & 10.89 & 49.08 & 10.89 & 48.16 & 3.26 & 8.85 & - & 52.86 \\
\raisebox{-0.22\height}{\includegraphics[width=0.02\textwidth]{close.png}}\raisebox{-0.22\height}{\includegraphics[width=0.02\textwidth]{chunk.png}} StandardRAG & 12.52 & 60.01 & 15.41 & 62.51 & 2.92 & 50.40 & 10.69 & 65.13 & 14.70 & 57.25 & 21.92 & 68.43 & 3.39 & 13.03 & 52.69 & 60.62 \\
\raisebox{-0.22\height}{\includegraphics[width=0.02\textwidth]{open.png}}\raisebox{-0.22\height}{\includegraphics[width=0.02\textwidth]{none.png}} SFT & 12.40 & 52.31 & 16.48 & 51.35 & 5.04 & 51.31 & 11.23 & 58.20 & 16.95 & 46.42 & 33.02 & 59.98 & 9.64 & 15.85 & - & 53.26 \\
\rowcolor{R1!15} \raisebox{-0.22\height}{\includegraphics[width=0.02\textwidth]{open.png}}\raisebox{-0.22\height}{\includegraphics[width=0.02\textwidth]{none.png}} R1 & 28.45 & 56.92 & 25.33 & 55.38 & 8.07 & 47.53 & 21.51 & 55.11 & 27.11 & 48.65 & 47.91 & 60.74 & 19.66 & 26.40 & - & 54.06 \\
\rowcolor{Search-R1!15} \raisebox{-0.22\height}{\includegraphics[width=0.02\textwidth]{open.png}}\raisebox{-0.22\height}{\includegraphics[width=0.02\textwidth]{chunk.png}} Search-R1 & 38.04 & 54.39 & 43.84 & 69.32 & 7.65 & 46.43 & 37.96 & 52.90 & 38.67 & 63.74 & 47.99 & 60.37 & 28.65 & 35.69 & 49.99 & 57.86 \\
\rowcolor{R1-Searcher!15} \raisebox{-0.22\height}{\includegraphics[width=0.02\textwidth]{open.png}}\raisebox{-0.22\height}{\includegraphics[width=0.02\textwidth]{chunk.png}} R1-Searcher & 23.50 & 55.86 & 42.44 & 64.60 & 12.81 & 50.07 & 36.53 & 63.33 & 40.18 & 66.23 & 54.00 & 60.52 & 27.08 & 34.91 & 49.98 & 60.10 \\
\rowcolor{Graph-R1!15} \raisebox{-0.22\height}{\includegraphics[width=0.02\textwidth]{open.png}}\raisebox{-0.22\height}{\includegraphics[width=0.02\textwidth]{graph.png}} Graph-R1 (ours) & \textbf{57.56} & \textbf{76.45} & \textbf{56.75} & \textbf{77.46} & \textbf{40.51} & \textbf{67.84} & \textbf{44.75} & \textbf{69.92} & \textbf{45.65} & \textbf{71.27} & \textbf{62.31} & \textbf{75.01} & \textbf{42.45} & \textbf{51.26} & \textbf{60.19} & \textbf{72.99}\\
\midrule
\multicolumn{17}{c}{\textbf{\textit{Qwen2.5-7B-Instruct}}} \\
\raisebox{-0.22\height}{\includegraphics[width=0.02\textwidth]{close.png}}\raisebox{-0.22\height}{\includegraphics[width=0.02\textwidth]{none.png}} NaiveGeneration & 12.25 & 66.75 & 16.58 & 65.31 & 4.06 & 65.47 & 13.00 & 69.56 & 12.82 & 60.50 & 24.51 & 72.65 & 3.12 & 13.87 & - & 66.71 \\
\raisebox{-0.22\height}{\includegraphics[width=0.02\textwidth]{close.png}}\raisebox{-0.22\height}{\includegraphics[width=0.02\textwidth]{chunk.png}} StandardRAG & 12.75 & 60.06 & 21.10 & 66.13 & 4.53 & 59.84 & 15.97 & 70.49 & 16.10 & 60.86 & 24.90 & 73.71 & 5.34 & 15.89 & 52.67 & 65.18 \\
\raisebox{-0.22\height}{\includegraphics[width=0.02\textwidth]{open.png}}\raisebox{-0.22\height}{\includegraphics[width=0.02\textwidth]{none.png}} SFT & 20.28 & 63.85 & 27.59 & 65.65 & 10.02 & 63.50 & 19.02 & 68.19 & 27.93 & 56.31 & 39.21 & 70.25 & 15.57 & 24.01 & - & 64.63 \\
\rowcolor{R1!15} \raisebox{-0.22\height}{\includegraphics[width=0.02\textwidth]{open.png}}\raisebox{-0.22\height}{\includegraphics[width=0.02\textwidth]{none.png}} R1 & 30.99 & 59.19 & 37.05 & 60.12 & 14.53 & 49.39 & 28.45 & 57.63 & 30.35 & 53.38 & 57.33 & 66.73 & 25.91 & 33.12 & - & 57.74 \\
\rowcolor{Search-R1!15} \raisebox{-0.22\height}{\includegraphics[width=0.02\textwidth]{open.png}}\raisebox{-0.22\height}{\includegraphics[width=0.02\textwidth]{chunk.png}} Search-R1 & 41.29 & 70.26 & 50.85 & 73.85 & 22.35 & 57.68 & 45.88 & 67.58 & 50.76 & 66.08 & 65.98 & 76.15 & 38.54 & 46.19 & 51.60 & 68.60 \\
\rowcolor{R1-Searcher!15} \raisebox{-0.22\height}{\includegraphics[width=0.02\textwidth]{open.png}}\raisebox{-0.22\height}{\includegraphics[width=0.02\textwidth]{chunk.png}} R1-Searcher & 33.96 & 69.61 & 46.36 & 74.56 & 16.63 & 59.05 & 44.93 & 68.54 & 47.12 & 66.74 & 64.76 & 75.95 & 34.51 & 42.29 & 51.26 & 69.08 \\
\rowcolor{Graph-R1!15} \raisebox{-0.22\height}{\includegraphics[width=0.02\textwidth]{open.png}}\raisebox{-0.22\height}{\includegraphics[width=0.02\textwidth]{graph.png}} Graph-R1 (ours) & \textbf{65.04} & \textbf{82.42} & \textbf{62.69} & \textbf{80.03} & \textbf{46.17} & \textbf{71.42} & \textbf{49.87} & \textbf{70.97} & \textbf{51.22} & \textbf{73.43} & \textbf{71.93} & \textbf{79.11} & \textbf{48.57} & \textbf{57.82} & \textbf{60.40} & \textbf{76.23}\\
\bottomrule
\end{tabular}%
}

\end{table*}

\textbf{End-to-end RL Objective $\mathcal{J}_{\text{GRPO}}(\theta)$. }
Given a dataset question $q \in \mathcal{D}_Q$, the agent interacts with the knowledge hypergraph $\mathcal{G}_H$ to generate a group of multi-turn reasoning trajectories $\{\tau_i\}_{i=1}^N \subseteq \mathcal{T}_q$, where each $\tau_i = ((\mathbf{s}_1^{(i)}, \mathbf{a}_1^{(i)}), \dots, (\mathbf{s}_T^{(i)}, \mathbf{a}_T^{(i)}))$ denotes a sequence of state-action pairs sampled from the environment. We optimize the policy $\pi_\theta$ using the GRPO-based objective:
\begin{equation}
\small
\begin{aligned}
&\mathcal{J}_{\text{GRPO}}(\theta) = \mathbb{E}_{\left[\mathbf{s}_1 \sim \{\mathbb{P}(q) \mid q \in \mathcal{D}_Q\},\; \{\tau_i\}_{i=1}^N \sim \pi_{\theta_{\text{old}}}(\mathcal{T}_q \mid \mathbf{s}_1;\mathcal{G}_H)\right]} \\
&\left[\!\frac{1}{N}\!\!\sum_{i=1}^N\!\frac{1}{|\tau_i|}\!\sum_{t=1}^{|\tau_i|}\!\min\!\left(\!\rho_{\theta}(\mathbf{a}_t^{(i)})\!\hat{A}(\tau_i), \text{clip}\!\left(\!\rho_{\theta}(\mathbf{a}_t^{(i)}),\!1\!\pm\!\epsilon\!\right)\!\hat{A}(\tau_i)\!\right)\right. \\
&\left. - \beta \, \mathbb{D}_{\mathrm{KL}}(\pi_\theta \| \pi_{\text{ref}})\right], \ \text{where}\ \ \rho_{\theta}(\mathbf{a}_t^{(i)})\!=\!\!
\frac{\pi_\theta(\mathbf{a}_t^{(i)} \mid \mathbf{s}_{t-1}^{(i)};\mathcal{G}_H))}{\pi_{\theta_{\text{old}}}(\mathbf{a}_t^{(i)} \mid \mathbf{s}_{t-1}^{(i)};\mathcal{G}_H))},\\
&\text{and}\ \ \hat{A}(\tau_i) = 
\frac{R(\tau_i) - \mathrm{mean} \left( \{ R(\tau_j) \}_{j=1}^{N} \right)}
     {F_{\mathrm{norm}} \left( \{ R(\tau_j) \}_{j=1}^{N} \right)}.
\end{aligned}
\end{equation}
Here, $\pi_\theta$ is the current policy, and $\pi_{\theta_{\text{old}}}$ is the behavior policy used for sampling. The importance ratio $\rho_\theta(\mathbf{a}_t^{(i)})$ adjusts for distribution shift, while the advantage $\hat{A}(\tau_i)$ normalizes the reward using a scaling function $F_{\mathrm{norm}}(\cdot)$ (e.g., standard deviation). The $\text{clip}(\cdot)$ operator stabilizes updates by constraining policy shifts. A KL term $\mathrm{D}_{\mathrm{KL}}(\pi_\theta \,\|\, \pi_{\text{ref}})$ regularizes toward a reference policy $\pi_{\text{ref}}$, with $\beta$ controlling its strength. This objective encourages high-reward, stable reasoning over $\mathcal{G}_H$.

\textbf{Outcome-directed Reward Function $R(\tau)$. }
To meet outcome requirements, we define a reward function $R(\tau)$ composed of two parts: a \textit{format reward} $R_{\text{format}}(\tau)$ and an \textit{answer reward} $R_{\text{answer}}(\mathbf{a}_T^{\text{ans}})$, promoting both thoughtful retrieval and accurate answer generation.

\textit{(i) Format Reward. }
The format reward $R_{\text{format}}(\tau)$ encourages the agent to follow the intended reasoning structure. At each step $(\mathbf{s}_t, \mathbf{a}_t)$, we check whether the output includes a well-formed block $(\mathbf{a}_t^{\text{think}}, \alpha_t, \mathbf{a}_t^{\text{out}})$. Each valid step receives $0.5$ reward, capped at $1.0$ overall:
\begin{equation}
\small
R_{\text{format}}(\tau) = \min\left(1.0,\; 0.5 \cdot \sum_{t=1}^{T} \mathbb{I}\left\{(\mathbf{a}_t^{\text{think}}, \alpha_t, \mathbf{a}_t^{\text{out}}) \right\}\right),
\end{equation}
where $\mathbb{I}\{\cdot\}$ is an indicator function that returns $1$ if the step output matches the expected format.

\textit{(ii) Answer Reward. }
The answer reward $R_{\text{answer}}(\mathbf{a}_T^{\text{ans}})$ measures the semantic correctness of the generated answer $\mathbf{a}_T^{\text{ans}}$ by comparing it with the ground-truth answer $y_q^*$ using a token-level F1 score $R_{\text{answer}}(\mathbf{a}_T^{\text{ans}})$.

\textit{(iii) Overall Outcome Reward.}
The total reward for a reasoning trajectory $\tau$ is defined as:
\begin{equation}
\small
R(\tau) = -1.0 + R_{\text{format}}(\tau) + \mathbb{I}\{R_{\text{format}}(\tau) = 1.0\} \cdot R_{\text{answer}}(\mathbf{a}_T^{\text{ans}}),
\end{equation}
where $\mathbf{a}_T^{\text{ans}} \in \tau$, ensuring that answer correctness is only rewarded when the format is structurally valid.

\textbf{Proposition 3.} \textit{End-to-end RL bridges the gap between graph-based knowledge and language.}\vspace{-3mm}
\begin{proof}
We provide quantitative experimental results in Section~\ref{5.6} and qualitative proofs in Appendix~\ref{proof3}.
\end{proof}

\section{Experiments}
This section presents the experimental setup, main results, and analysis. We answer the following research questions (RQs):
\textbf{RQ1: }Does Graph-R1 outperform other methods?
\textbf{RQ2: }Does the main component of Graph-R1 work, and how is its comparative analysis?
\textbf{RQ3-5: }How are construction cost, retrieval efficiency, and generation quality?

\subsection{Experimental Setup}

\textbf{Datasets. }
To evaluate the performance of Graph-R1, we conduct experiments across six standard RAG datasets~\citep{FlashRAG}:
2WikiMultiHopQA (\textbf{2Wiki.})~\citep{2WikiMultiHopQA}, \textbf{HotpotQA}~\citep{HotpotQA}, \textbf{Musique}~\citep{Musique}, Natural Questions (\textbf{NQ})~\citep{NQ}, \textbf{PopQA}~\citep{PopQA}, and \textbf{TriviaQA}~\citep{TriviaQA}. More details are in Appendix~\ref{Dataset}.  

\textbf{Baselines. }
We mainly compare Graph-R1 with \textbf{NaiveGeneration}, \textbf{StandardRAG}~\citep{RAG}, \textbf{SFT}~\citep{SFT}, \textbf{R1}~\citep{GRPO}, \textbf{Search-R1}~\citep{Search-R1}, and \textbf{R1-Searcher}~\citep{R1-Searcher} at three \texttt{Qwen2.5}~\citep{Qwen2.5} scales: 1.5 B, 3 B, and 7 B. We also compare \textbf{GraphRAG}~\citep{GraphRAG}, \textbf{LightRAG}~\citep{LightRAG}, \textbf{PathRAG}~\citep{PathRAG}, \textbf{HippoRAG2}~\citep{HippoRAG2}, and \textbf{HyperGraphRAG}~\citep{HyperGraphRAG} based on \texttt{GPT-4o-mini}~\citep{GPT-4o} as a reference. More details are in Appendix~\ref{Baseline}.

% \textbf{Evaluation Metrics. }
% We evaluate Graph-R1 and baselines with four metrics: Exact Match (\textbf{EM}), \textbf{F1}, Retrieval Similarity (\textbf{R-S}), and Generation Evaluation (\textbf{G-E}). More details are in Appendix~\ref{Evaluation}.
\textbf{Evaluation Metrics. }
We evaluate Graph-R1 and baselines with four metrics: Exact Match (\textbf{EM}), \textbf{F1}, Retrieval Similarity (\textbf{R-S}), and Generation Evaluation (\textbf{G-E}), which together assess answer correctness, retrieval relevance, and overall generation quality. These metrics provide a comprehensive evaluation of both retrieval and generation components of the framework. More details are in Appendix~\ref{Evaluation}.

\textbf{Implementation Details. }
We use \texttt{GPT-4o-mini} for knowledge construction in Graph-R1 and GraphRAG baselines. For retrieval, we use \texttt{bge-large-en-v1.5}~\citep{BAAIembedding} in all variants. All experiments are done on 4 NVIDIA A100 GPUs (80GB) with identical training and inference settings. More details are in Appendix~\ref{Implementation}.

\begin{figure*}[t]
\footnotesize
\centering

% -------- Row 1: (a) + (b) --------
\begin{subfigure}[t]{0.55\textwidth}
  \centering
  \vspace{-37.5mm} % 你原来的偏移保留；如不需要可删
  \fontsize{6.5pt}{6.7pt}\selectfont
  \setlength{\tabcolsep}{0.9mm}
  \begin{tabular}{l cccc cccc|cccc}
  \toprule
  \multirow{2.5}{*}{\textbf{Method}} & \multicolumn{4}{c}{\textbf{2Wiki.}} & \multicolumn{4}{c}{\textbf{HotpotQA}} & \multicolumn{4}{c}{\textbf{Avg.}} \\
  \cmidrule(lr){2-5} \cmidrule(lr){6-9} \cmidrule(lr){10-13}
   & \textbf{EM} & \textbf{F1} & \textbf{R-S} & \textbf{G-E} & \textbf{EM} & \textbf{F1} & \textbf{R-S} & \textbf{G-E} & \textbf{EM} & \textbf{F1} & \textbf{R-S} & \textbf{G-E}\\
  \midrule
  \multicolumn{13}{c}{\textbf{\textit{Qwen2.5-3B-Instruct}}} \\
  \rowcolor{Graph-R1!15} Graph-R1 & \textbf{50.00} & \textbf{57.56} & \textbf{55.78} & \textbf{76.45} & \textbf{50.78} & \textbf{56.75} & \textbf{54.74} & \textbf{77.46} & \textbf{50.39} & \textbf{57.16} & \textbf{55.26} & \textbf{76.96}\\
  \textit{w/o K.C.} & 36.33 & 44.94 & 53.46 & 65.83 & 40.63 & 47.27 & 53.23 & 72.69 & 38.48 & 46.11 & 53.35 & 69.26\\
  \textit{w/o M.I.} & 21.88 & 34.34 & 54.70 & 66.59 & 30.86 & 37.64 & 53.34 & 64.75 & 26.37 & 35.99 & 54.02 & 65.67\\
  \textit{w/o R.L.} & 0.78 & 8.91 & 10.16 & 47.14 & 5.47 & 12.56 & 14.44 & 58.60 & 3.13 & 10.74 & 12.30 & 52.87\\
  \midrule
  \multicolumn{13}{c}{\textbf{\textit{Qwen2.5-7B-Instruct}}} \\
  \rowcolor{Graph-R1!15} Graph-R1 & \textbf{55.47} & \textbf{65.04} & \textbf{55.24} & \textbf{82.42} & \textbf{57.03} & \textbf{62.69} & \textbf{56.27} & \textbf{80.03} & \textbf{56.25} & \textbf{63.87} & \textbf{55.76} & \textbf{81.23}\\
  \textit{w/o K.C.} & 44.14 & 51.81 & 54.10 & 75.90 & 49.22 & 55.93 & 54.14 & 76.78 & 46.68 & 53.87 & 54.12 & 76.34\\
  \textit{w/o M.I.} & 37.50 & 44.78 & 54.54 & 69.98 & 40.63 & 47.04 & 54.58 & 69.63 & 39.07 & 45.91 & 54.56 & 69.81\\
  \textit{w/o R.L.} & 0.00 & 18.25 & 54.63 & 75.81 & 3.12 & 17.33 & 53.80 & 78.92 & 1.56 & 17.79 & 54.22 & 77.37\\
  \bottomrule
  \end{tabular}
  \caption{Ablation Study}
  \label{F5a}
\end{subfigure}
\hspace{4mm}
\begin{subfigure}[t]{0.3\textwidth}
  \centering
  \includegraphics[width=\linewidth]{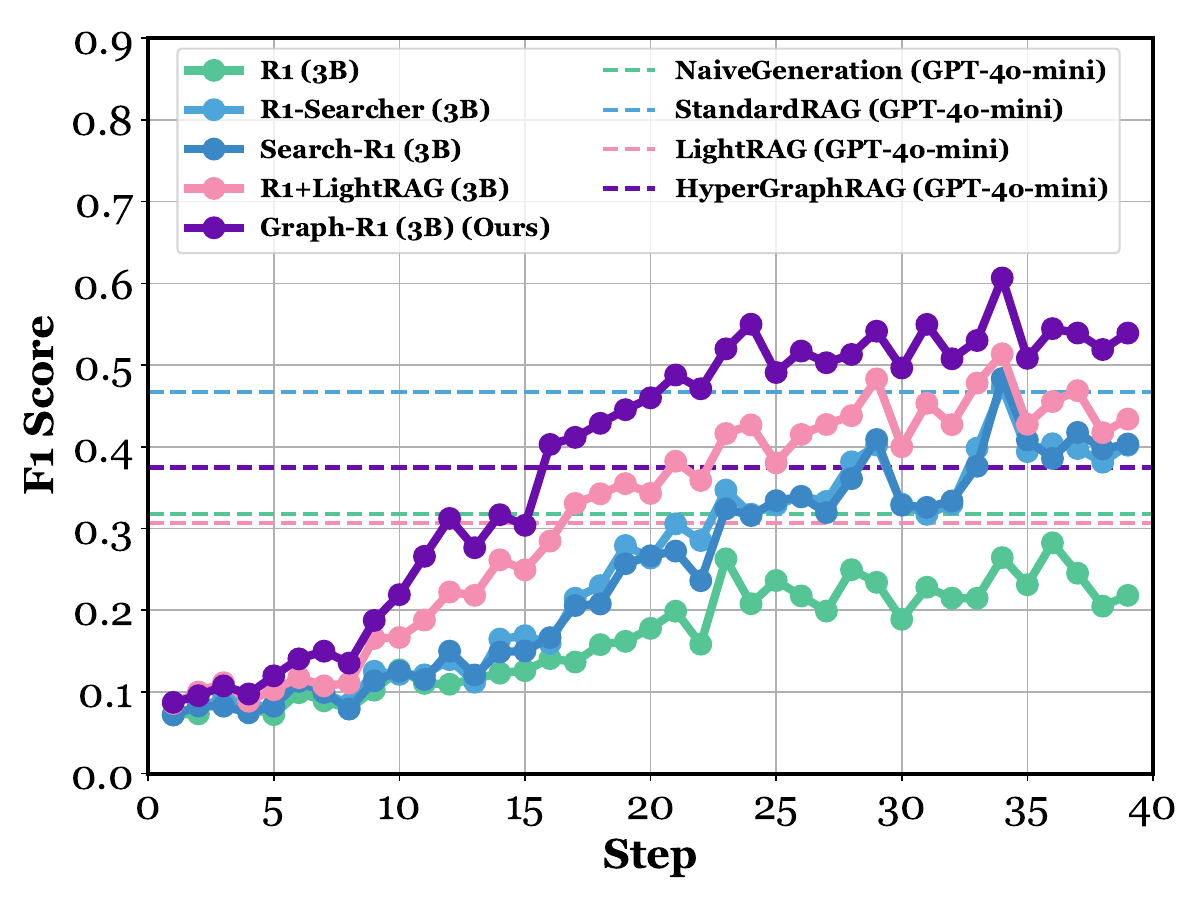}
  \vspace{-5.5mm}
  \caption{Representations}
  \label{F5b}
\end{subfigure}

% \vspace{-1mm}

% -------- Row 2: (c)-(f) --------
\begin{subfigure}[t]{0.24\textwidth}
  \centering
  \includegraphics[width=\linewidth]{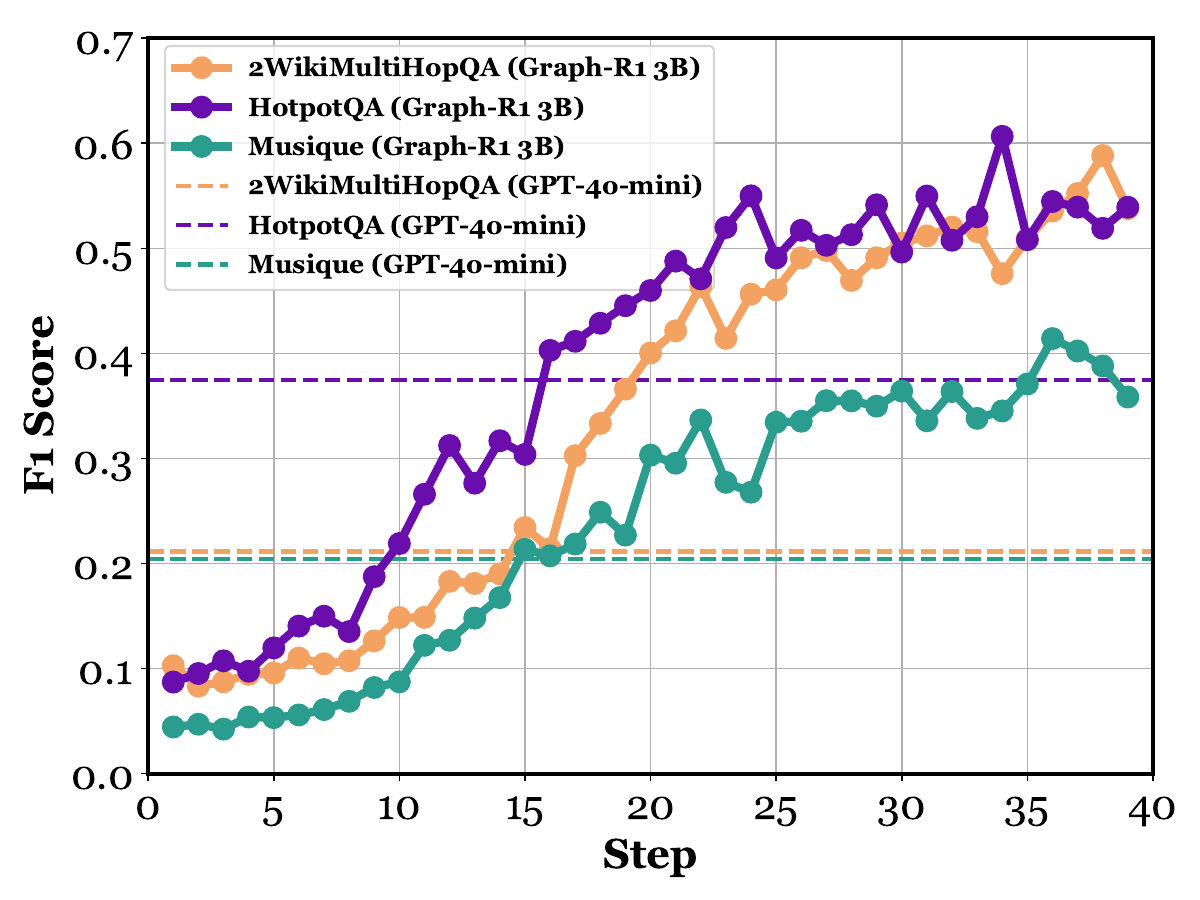}
  \vspace{-5mm}
  \caption{Datasets}
  \label{F5c}
\end{subfigure}\hspace{-1mm}
\begin{subfigure}[t]{0.24\textwidth}
  \centering
  \includegraphics[width=\linewidth]{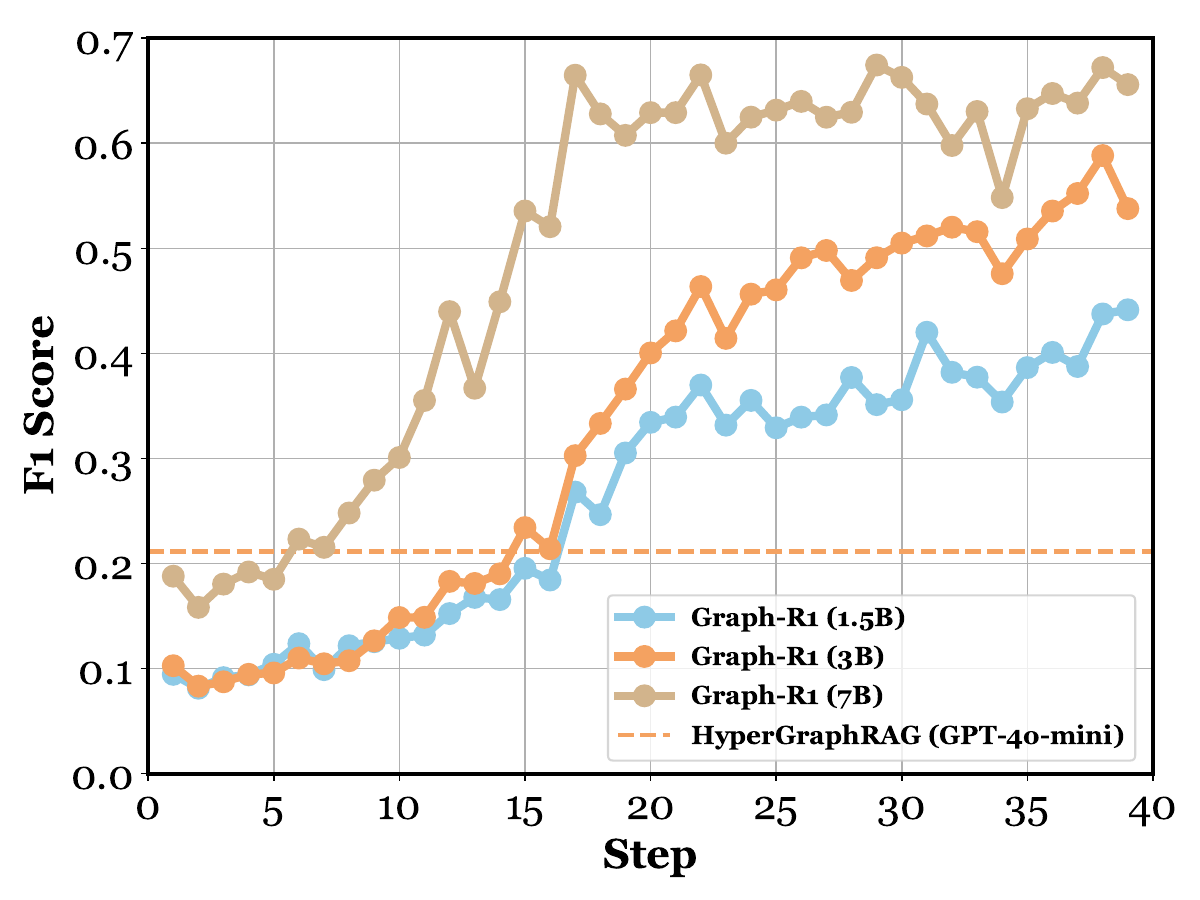}
  \vspace{-5mm}
  \caption{Parameters}
  \label{F5d}
\end{subfigure}\hspace{-1mm}
\begin{subfigure}[t]{0.24\textwidth}
  \centering
  \includegraphics[width=\linewidth]{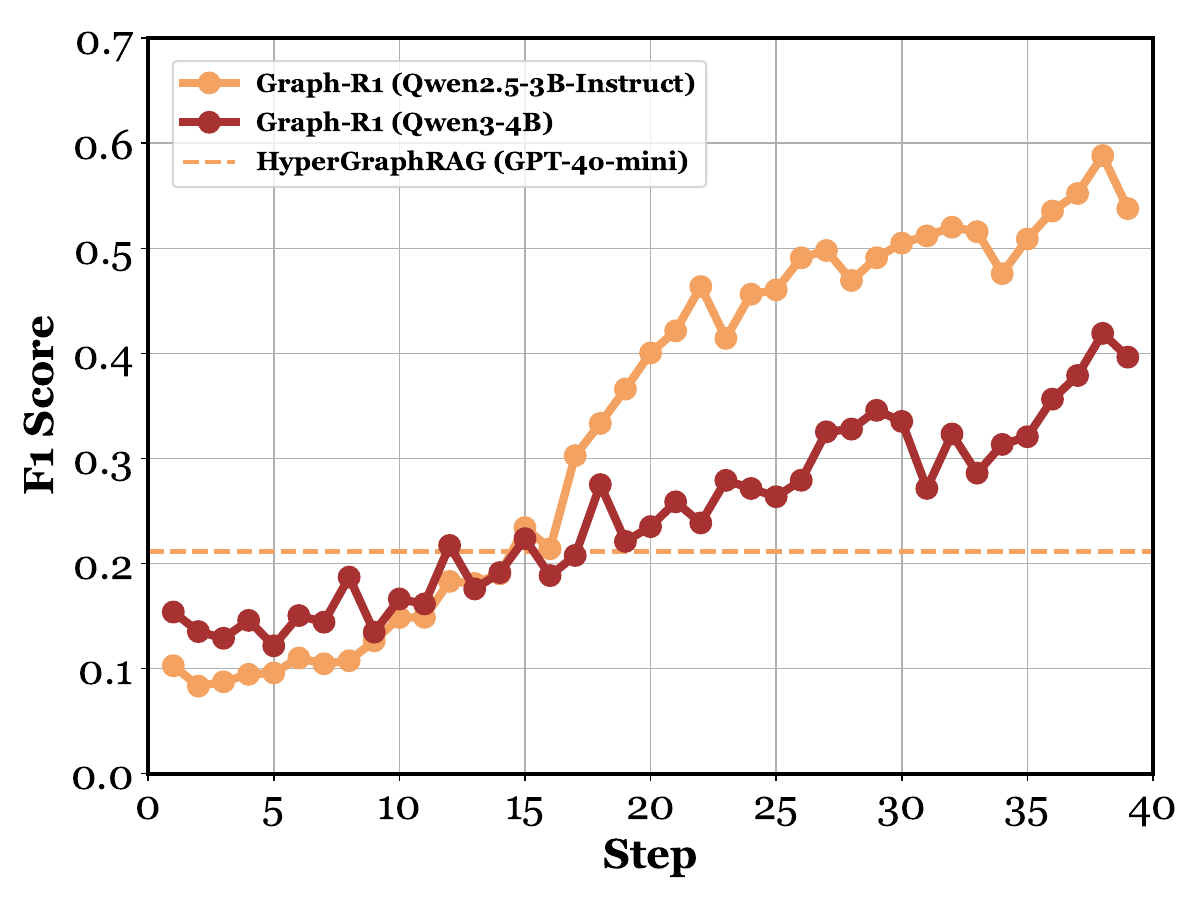}
  \vspace{-5mm}
  \caption{Qwen3}
  \label{F5e}
\end{subfigure}\hspace{-1mm}
\begin{subfigure}[t]{0.24\textwidth}
  \centering
  \includegraphics[width=\linewidth]{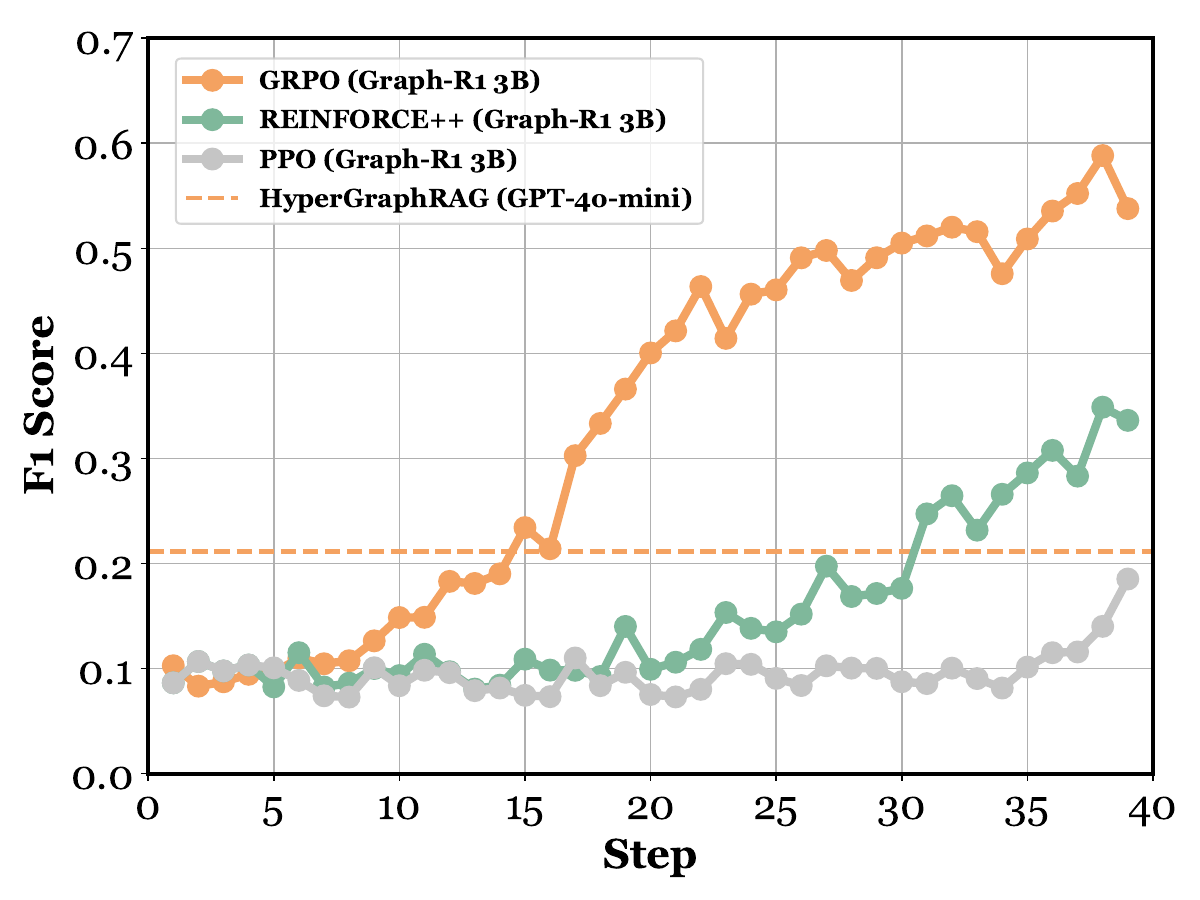}
  \vspace{-5mm}
  \caption{Algorithms}
  \label{F5f}
\end{subfigure}

\vspace{-1mm}
\caption{(a) Ablation study of Graph-R1. (b--f) Performance comparison across different kinds of knowledge representations, RAG datasets, model parameters, Qwen versions, and RL algorithms.}
\label{F5}
\end{figure*}

\subsection{Main Results (RQ1)}
\label{5.2}
As shown in Table~\ref{T2}, we compare Graph-R1 with baselines across different base models, and observe that Graph-R1 consistently outperforms all baselines. In addition, we have two key observations.

\textbf{RL Unlocks the Power of Graph Representations. }
Prompt-only GraphRAG methods often underperform StandardRAG, showing that graph structures alone are not sufficient. Graph-R1, with multi-turn RL optimization, fully exploits structural signals, achieving 57.28 F1 under Qwen2.5-7B-Instruct, surpassing StandardRAG (32.05), HyperGraphRAG (29.40) and Search-R1 (46.19).

\textbf{Larger Base Model Further Enhances Performance. }
As base model size increases from 1.5B to 3B and 7B, Graph-R1 achieves steadily higher F1 scores: 40.09, 51.26, and 57.82. Moreover, its gap over other RL-enhanced baselines such as Search-R1 and R1-Searcher becomes increasingly evident. This shows that larger models better exploit the synergy between graph structures and RL.

\subsection{Ablation Study and Comparative Analysis (RQ2)}
Figures~\ref{F5} shows an ablation study and comparative analysis.

\textbf{Ablation Study.}
We remove three core components of Graph-R1: knowledge construction (K.C.), multi-turn interaction (M.I.), and reinforcement learning (R.L.), to assess their individual contributions. As shown in Figure~\ref{F5a}, removing any module leads to performance degradation.

\textbf{Comparison with Different Knowledge Representations.}
As shown in Figures~\ref{F4} and~\ref{F5b}, models without external knowledge (green) perform the worst. Chunk-based knowledge with RL (blue) performs better, but is still inferior to graph-based methods using binary relations (pink), while hypergraph-based knowledge with RL (red) achieves the highest ceiling. This demonstrates that, when combined with RL, stronger knowledge representations yield higher performance potential.

\textbf{Comparison across Datasets and Base Models.}
As shown in Figures~\ref{F5c} and~\ref{F5d}, Graph-R1 consistently outperforms baselines across different datasets and parameter sizes, showcasing strong scalability. Interestingly, Figure~\ref{F5e} shows that when Graph-R1 is trained on Qwen3 (4B)~\citep{Qwen3}, which is already well trained by RL, the model tends to over-rely on its own internal reasoning. Despite a stronger starting point, its overall performance ceiling appears slightly lower.

\textbf{Comparison with Different RL Algorithms.}
Figure~\ref{F5f} compares different RL strategies. GRPO significantly outperforms REINFORCE++~\citep{RPP} and PPO~\citep{PPO}, achieving the highest F1. This confirms that GRPO facilitates more stable training and stronger multi-turn graph reasoning, making it a favorable choice for training agentic GraphRAG models.

\begin{figure*}[t]
\footnotesize
\centering

% -------- Row 1: (a) + (b) --------
\begin{subfigure}[t]{0.55\textwidth}
\vspace{-40mm}
\centering
    \fontsize{7.7pt}{9.3pt}\selectfont
    \setlength{\tabcolsep}{0.9mm}{
    \begin{tabular}{lcc|cc|cc|c}
        \toprule
        \multirow{2.5}{*}{\textbf{Method}} & \multicolumn{4}{c}{\textbf{Knowledge Construction}} & \multicolumn{3}{c}{\textbf{Retrieval \& Generation}} \\
        \cmidrule(lr){2-5} \cmidrule(lr){6-8}
        & \textbf{TP1KT} & \textbf{CP1MT} & \textbf{\#Node} & \textbf{\#Edge} & \textbf{TPQ} & \textbf{CP1KQ} & \textbf{F1} \\
        \midrule
        NaiveGeneration & 0 s & 0 \$ & - & - & 3.7 s & 0.16 \$ & 17.0 \\
        StandardRAG & 0 s & 0 \$ & - & - & 4.1 s & 1.35 \$ & 22.3 \\
        GraphRAG & 8.04 s & 3.35 \$ & 7,771 & 4,863 & 7.4 s & 3.97 \$ & 16.0 \\
        LightRAG & 6.84 s & 4.07 \$ & 59,197 & 24,596 & 12.2 s & 8.11 \$ & 16.6 \\
        PathRAG & 6.84 s & 4.07 \$ & 59,197 & 24,596 & 15.8 s & 8.28 \$ & 12.4 \\
        HippoRAG2 & 3.25 s & 1.26 \$ & 11,819 & 40,654 & 8.8 s & 7.68 \$ & 16.3 \\
        HyperGraphRAG & 6.76 s & 4.14 \$ & 173,575 & 114,426 & 9.6 s & 8.76 \$ & 21.1 \\
        \rowcolor{Graph-R1!15} Graph-R1 (7B) (ours) & 5.69 s & 2.81 \$ & 120,499 & 98,073 & 7.0 s & 0 \$ & 65.0 \\
        \bottomrule
    \end{tabular}}
  \caption{}
  \label{T3}
\end{subfigure}
\hspace{8mm}
\begin{subfigure}[t]{0.26\textwidth}
  \centering
  \includegraphics[width=\linewidth]{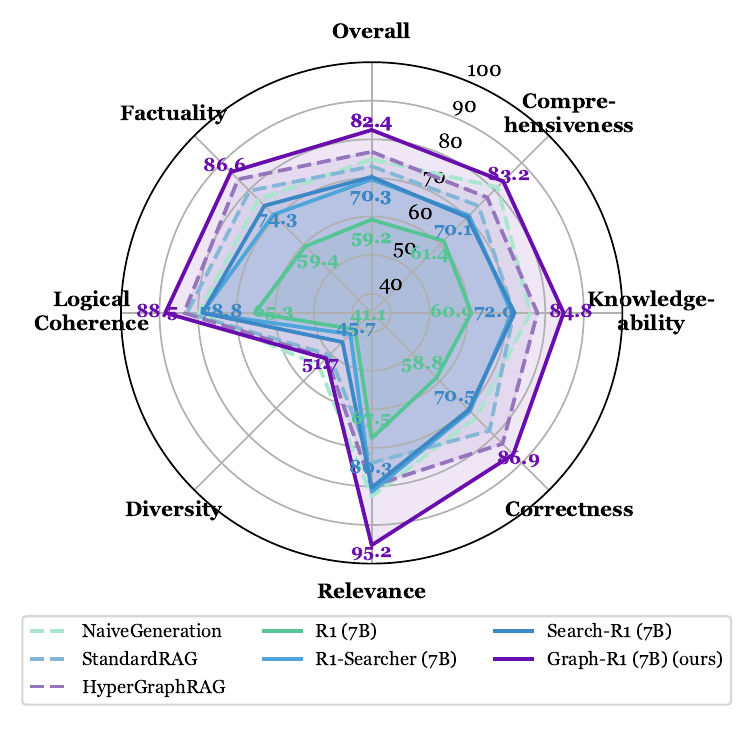}
  \vspace{-5.5mm}
  \caption{}
  \label{F7}
\end{subfigure}

\vspace{-1mm}
\caption{(a) Time \& Cost Comparisons on 2Wiki. (b) Generation Evaluations.}
\label{F5}
\end{figure*}

\begin{figure*}[t]
\footnotesize
\centering

\begin{subfigure}{0.295\textwidth}
  \includegraphics[width=\linewidth]{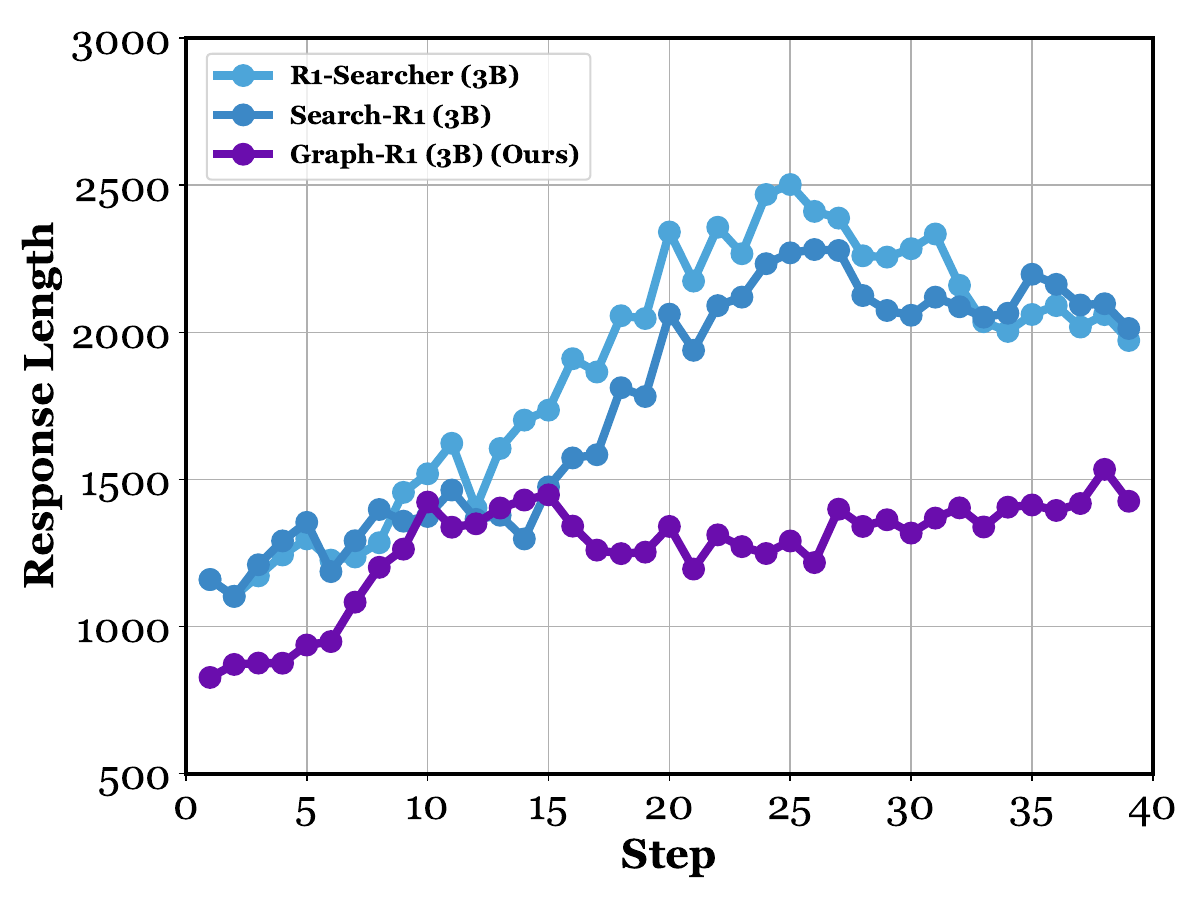}
  \vspace{-4.5mm}
  \caption{Response Length}
  \label{F6a}
\end{subfigure}
\hspace{4mm}
\begin{subfigure}{0.295\textwidth}
  \includegraphics[width=\linewidth]{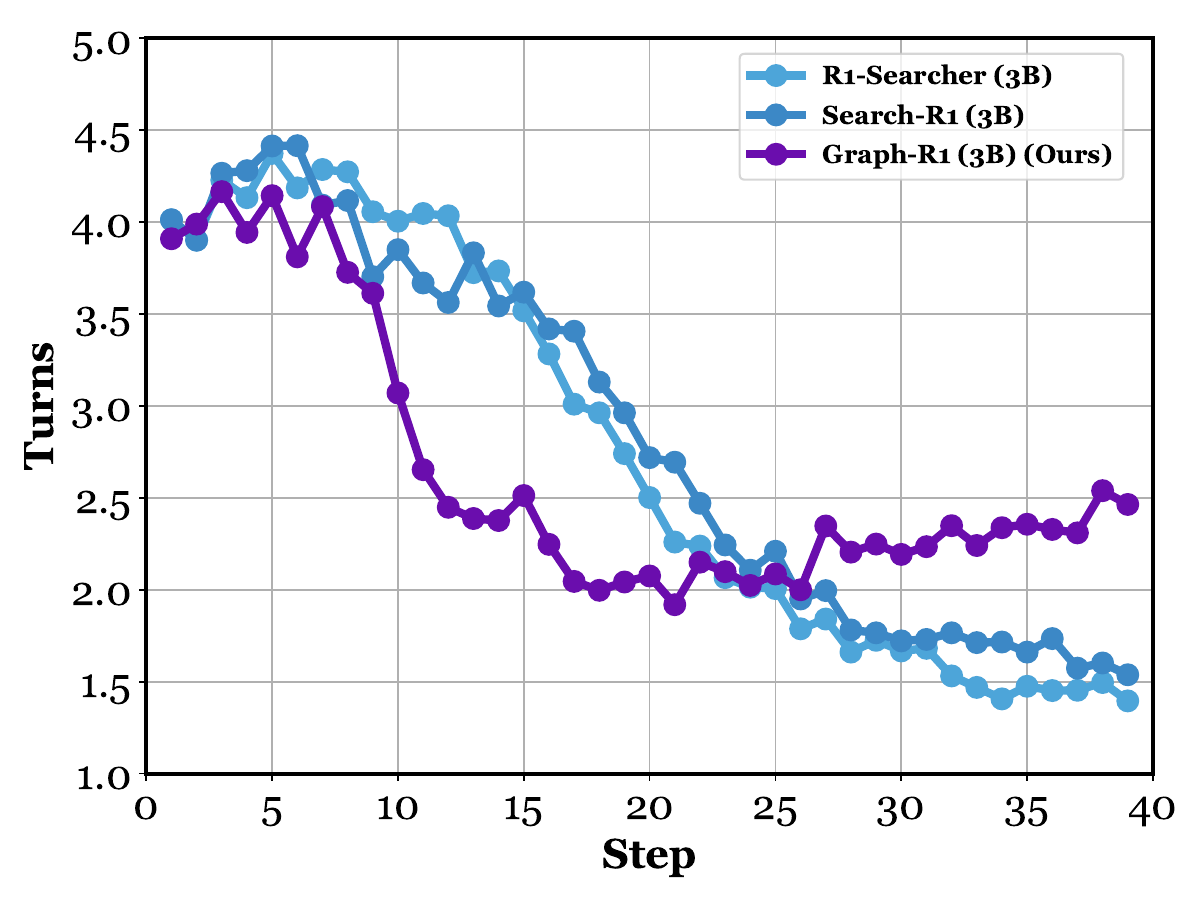}
  \vspace{-4.5mm}
  \caption{Turns of Interaction}
  \label{F6b}
\end{subfigure}
\hspace{4mm}
\begin{subfigure}{0.295\textwidth}
  \includegraphics[width=\linewidth]{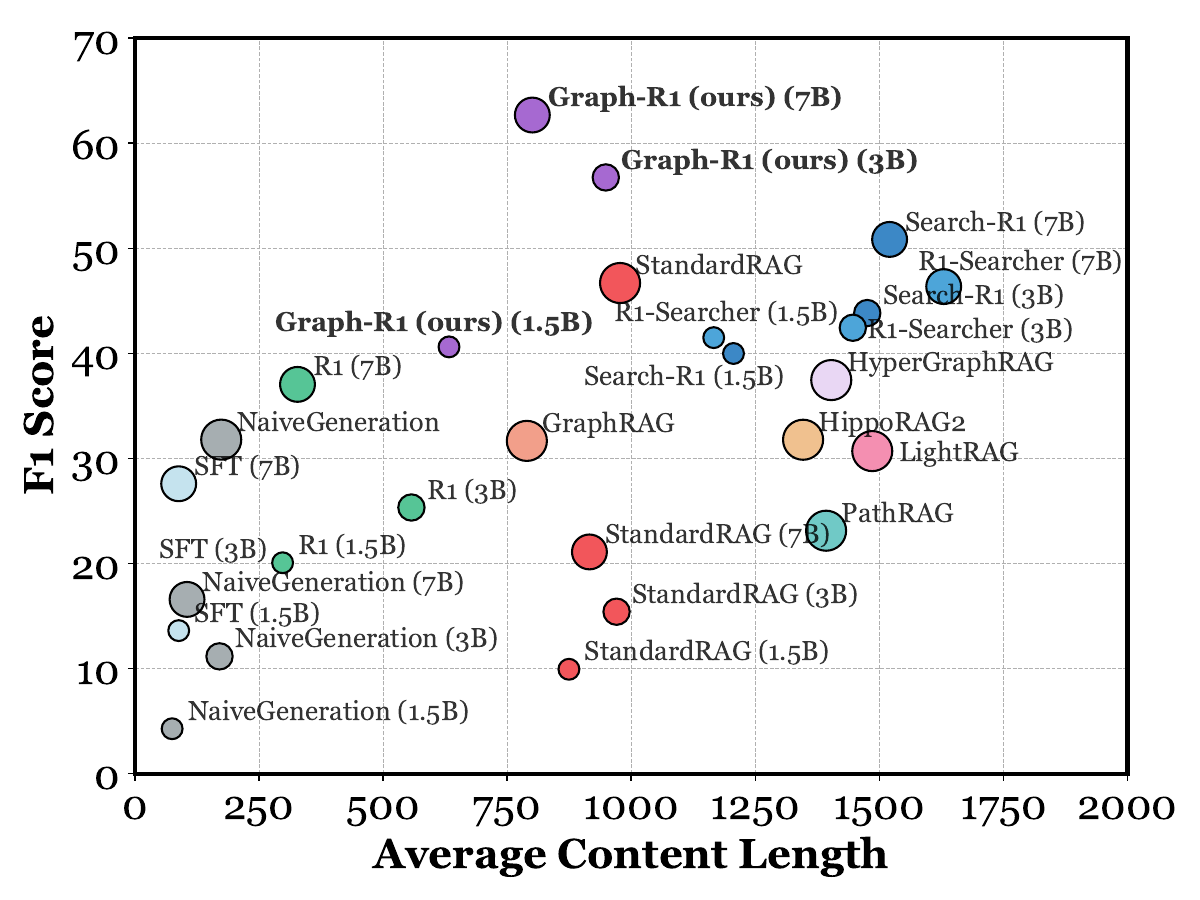}
  \vspace{-4.5mm}
  \caption{Efficiency Comparison}
  \label{F6c}
\end{subfigure}

\vspace{-1mm}
\caption{Step-wise response length \& turns of interaction, and efficiency comparison on HotpotQA.}
\label{F6}
\end{figure*}

\subsection{Analysis of Graph-R1's Construction Cost (RQ3)}
\label{5.4}
As shown in Table~\ref{T3}, we utilize metrics: time per 1K tokens (TP1KT), cost per 1M tokens (CP1MT), number of nodes \& edges, time per query (TPQ), cost per 1K queries (CP1KQ), and final F1 score.

\textbf{Construction Cost. }
Graph-R1 requires only 5.69 seconds and \$2.81 per 1K tokens for knowledge construction, lower than GraphRAG (8.04s, \$3.35) and HyperGraphRAG (6.76s, \$4.14). Generating over 120K nodes and 98K edges, Graph-R1 maintains a semantically rich structure.

\textbf{Generation Cost. }
By leveraging end-to-end RL and localized knowledge retrieval, Graph-R1 achieves not only the best F1 but also a response time of 7.0s per query and a generation cost of \$0, outperforming baselines such as HyperGraphRAG (9.6s, \$8.76), highlighting its superior potential for real-world deployment.

\subsection{Analysis of Graph-R1's Retrieval Efficiency (RQ4)}
\label{5.5}
As shown in Figure~\ref{F6}, to evaluate Graph-R1's retrieval efficiency, we analyze it from (a) response length, (b) number of interaction turns, and (c) performance with average retrieval content lengths.

\textbf{Tendency toward More Concise Thinking and Adequate Interaction. }
As shown in Figures~\ref{F6a} and~\ref{F6b}, Graph-R1 generates shorter responses and conducts more interaction turns, averaging around 1200-1500 tokens and 2.3-2.5 turns, leading to more stable and accurate retrieval.

\textbf{Balancing Performance and Retrieved Content Length. }
As shown in Figure~\ref{F6c}, Graph-R1 achieves the highest F1 scores with a moderate amount of average retrieved content compared to other methods, balancing input length and performance through its multi-turn interaction strategy.

\subsection{Analysis of Graph-R1's Generation Quality (RQ5)}
\label{5.6}
As shown in Figure~\ref{F7}, we evaluate the generation quality in seven dimensions and present a case study in Table~\ref{T4}.

\textbf{High-Quality Generation Performance. }
Graph-R1 outperforms all RL-based baselines and achieves generation quality comparable to GPT-4o-mini-based methods like HyperGraphRAG, with strong results in Correctness (86.9), Relevance (95.2), and Logical Coherence (88.5).

\textbf{RL Bridges the Gap Between Graph \& Language. }
HyperGraphRAG performs similarly to StandardRAG, indicating limited gains from graph structure alone. In contrast, Graph-R1 achieves a much higher Overall score (82.4 vs. 70.3) than Search-R1, showing that graph-based reasoning becomes truly effective when combined with RL.

\section{Conclusion}
In this work, we introduce Graph-R1, the agentic GraphRAG framework powered by end-to-end RL. By introducing lightweight knowledge hypergraph construction and modeling retrieval as a multi-turn interaction process, Graph-R1 bridges graph-structured knowledge with natural language generation. A unified reward mechanism enables outcome-directed reasoning. Experiments across six benchmarks demonstrate Graph-R1’s superiority in accuracy, retrieval efficiency, and generation quality.

\clearpage

\section*{Acknowledgments}
This work is supported by the National Natural Science Foundation of China (Grant No. 62473271, Grant No. 62176026, and Grant No. 62406036) and the Fundamental Research Funds for the Beijing University of Posts and Telecommunications (Grant No. 2025AI4S03). We also acknowledge the support from the Engineering Research Center of Information Networks, Ministry of Education, China. This work is also supported by the National Research Foundation, Singapore, under its Frontier Competitive Research Programme (NRF-F-CRP-2024-0005). Any opinions, findings and conclusions or recommendations expressed in this material are those of the authors and do not reflect the views of National Research Foundation, Singapore.

\section*{Impact Statement}
% This work introduces Graph-R1, a novel agentic GraphRAG framework that formulates graph-based retrieval and reasoning as a multi-turn interaction process optimized via end-to-end reinforcement learning. While effective, it still faces several limitations, including the efficiency and stability of reinforcement learning over long reasoning trajectories and the scalability of learned retrieval policies to complex graph structures. To address these challenges, we outline four future directions: (1) improving reinforcement learning efficiency and credit assignment for multi-step reasoning in graph environments, (2) enhancing adaptive exploration and pruning strategies over large and densely connected knowledge hypergraphs, (3) extending the framework to specialized domains such as medicine and law, where structured multi-entity relations are critical but costly to construct, and (4) expanding Graph-R1 to multimodal and multilingual settings to support reasoning over heterogeneous knowledge sources. 
This work poses no significant ethical concerns, as it relies solely on publicly available datasets, and its broader societal impact is expected to be positive by advancing reliable and structured knowledge-grounded reasoning for retrieval-augmented generation systems.
\nocite{langley00}

\bibliography{example_paper}
\bibliographystyle{icml2026}

%%%%%%%%%%%%%%%%%%%%%%%%%%%%%%%%%%%%%%%%%%%%%%%%%%%%%%%%%%%%%%%%%%%%%%%%%%%%%%%
%%%%%%%%%%%%%%%%%%%%%%%%%%%%%%%%%%%%%%%%%%%%%%%%%%%%%%%%%%%%%%%%%%%%%%%%%%%%%%%
% APPENDIX
%%%%%%%%%%%%%%%%%%%%%%%%%%%%%%%%%%%%%%%%%%%%%%%%%%%%%%%%%%%%%%%%%%%%%%%%%%%%%%%
%%%%%%%%%%%%%%%%%%%%%%%%%%%%%%%%%%%%%%%%%%%%%%%%%%%%%%%%%%%%%%%%%%%%%%%%%%%%%%%
\newpage
\appendix
\onecolumn
\section*{Appendix} 
\section{Prompts Used in Graph-R1}

\subsection{Knowledge HyperGraph Construction Prompt}
\label{AppendixA1}
\begin{figure}[h]
\vspace{-1mm}
\centering
\includegraphics[width=0.9\linewidth]{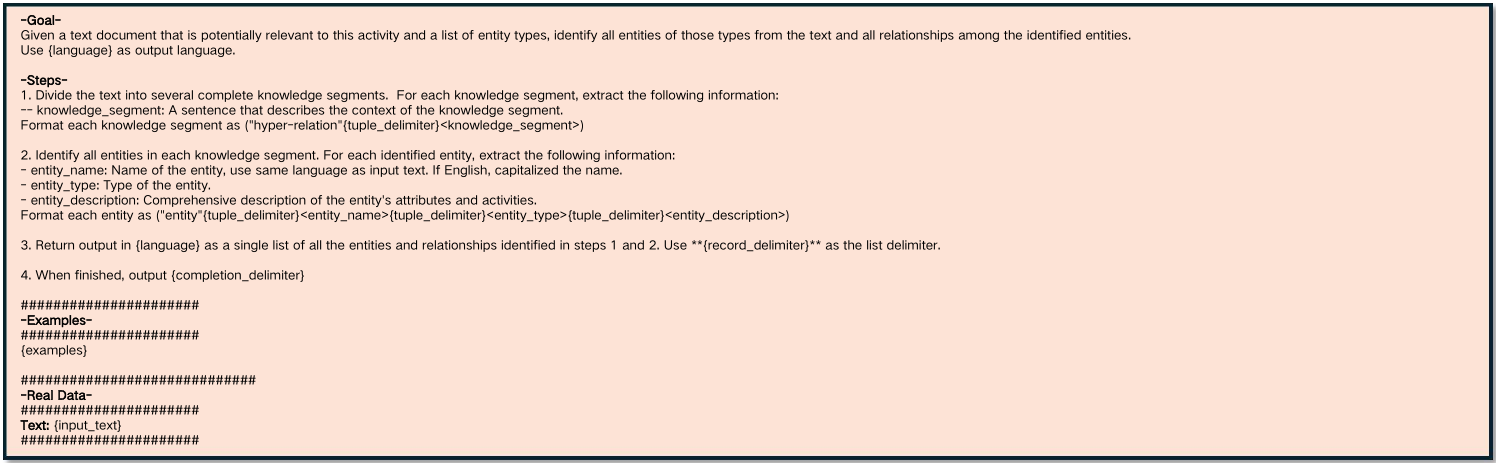}
\vspace{-1.5mm}
\caption{\label{prompt1}
Prompt for n-ary relation extraction $\pi_{\text{ext}}$ in Equation~\ref{E4}.}
\vspace{-2mm}
\end{figure}

As shown in Figure~\ref{prompt1}, we use the same n-ary relation-extraction prompt as HyperGraphRAG~\citep{HyperGraphRAG}. Moreover, we streamline knowledge-hypergraph construction by skipping confidence-score calculations and adopting a simpler semantic-retrieval method, reducing construction costs while maintaining equivalent knowledge-representation.

\subsection{Agentic Knowledge Reasoning Prompt}
\label{AppendixA2}
\begin{figure}[h]
\vspace{-1mm}
\centering
\includegraphics[width=0.9\linewidth]{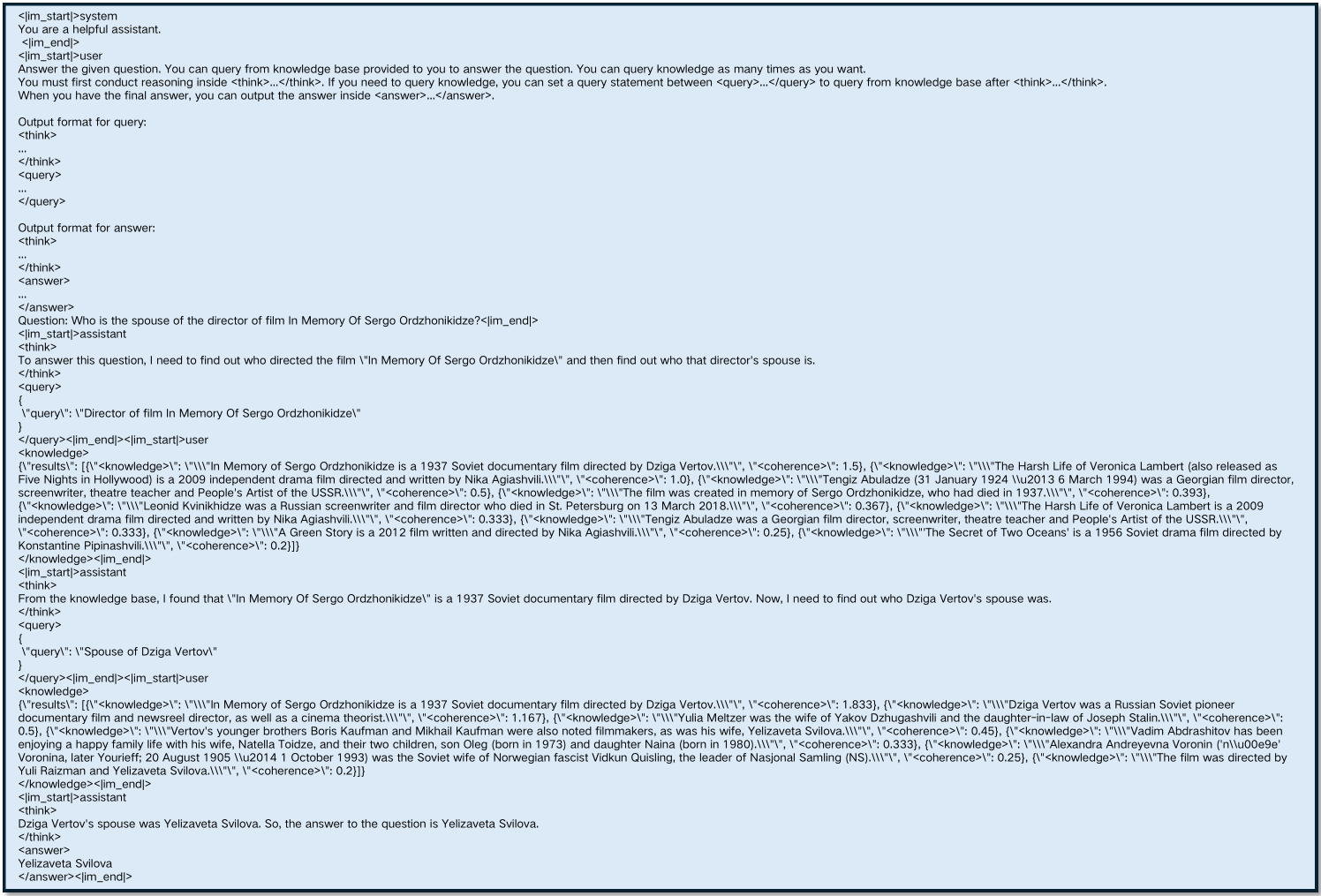}
\vspace{-1.5mm}
\caption{\label{prompt2}
Prompt for agentic knowledge reasoning $\pi_{\theta}$ in Equation~\ref{E6}.}
\vspace{-1.5mm}
\end{figure}

The initial knowledge-reasoning prompt has been shown in Table~\ref{T1}. Through several rounds of interaction with the environment, the Graph-R1 agent keeps adding information, readying the prompt for the next exchange. As shown in Figure~\ref{prompt2}, we present a case of the final prompt produced by the full agentic knowledge-reasoning process.

\newpage
\section{Theoretical Proof}
\label{proofs}
\subsection{Proof of Proposition 1}
\label{proof1}
\textbf{Proposition 1.}
\textit{Graph-structured knowledge boosts agent accuracy by richer representation.}\vspace{-2mm}
\begin{proof}
Let the knowledge base $\mathcal K$ be encoded into two forms: a graph $R_G$ and a linear chunk set $R_C$, where $R_C = g(R_G)$ is a deterministic transformation that discards edge information. For a query $Q$ and ground-truth answer $A^\star$, the agent’s internal belief at step $t$ is $h_t$. Each step performs retrieval $\mathcal E_t(R)$ and Bayesian update, forming the recurrence:
\begin{equation}
h_{t+1} = f(h_t, R).
\end{equation}
Define the Lyapunov function as $V_R(h_t) = -\log P(A^\star \mid h_t)$, measuring how far the agent is from certainty. Its update is:
\begin{equation}
\Delta V_R(h_t) = -\log \frac{P(\mathcal E_t(R) \mid A^\star)}{\sum_{a} P(a \mid h_t) P(\mathcal E_t(R) \mid a)}.
\end{equation}
Graphs can capture more relevant facts in shorter contexts due to explicit edges, leading to higher information density $\delta_R$ and more negative $\Delta V_R(h_t)$ in expectation. Thus, $V_R(h_t)$ decreases faster with graphs, indicating faster convergence. From an information-theoretic view, the mutual information evolves as:
\begin{equation}
I(A^\star; h_{t+1} \mid Q) = I(A^\star; h_t \mid Q) + I(A^\star; \mathcal E_t(R) \mid h_t, Q),
\end{equation}
and since graphs provide denser evidence, we have $I_{R_G}(A^\star; h_T \mid Q) \ge I_{R_C}(A^\star; h_T \mid Q)$.  
Then by Fano’s inequality,
\begin{equation}
P_e(R) \le \frac{H(A^\star \mid Q) - I_R(A^\star; h_T \mid Q) + 1}{\log |\mathcal A|},
\end{equation}
which implies $P_e(R_G) \le P_e(R_C)$, i.e., $\operatorname{Acc}(R_G) \ge \operatorname{Acc}(R_C)$, with strict inequality when the graph contains structural relations not recoverable from text.

In summary, the graph-structured representation offers higher information density per retrieval, accelerates belief convergence via Lyapunov descent, and accumulates more mutual information, leading to provably higher answer accuracy.
\end{proof}

\subsection{Proof of Proposition 2}
\label{proof2}
\textbf{Proposition 2.}
\textit{Multi-turn interaction with the graph environment improves retrieval efficiency.}\vspace{-2mm}
\begin{proof}
Let the graph-structured knowledge base be denoted by $R_G$, and let $A^\star$ be the ground-truth answer. Suppose the retrieval cost is measured by the number of tokens retrieved, and we fix a total budget of $B$ tokens. A single-turn retrieval strategy selects a fixed token set $\mathcal E_{\mathrm{static}}$ of size $B$, independent of any intermediate reasoning. This leads to a posterior belief $P(A^\star \mid Q, \mathcal E_{\mathrm{static}})$ and yields total information gain
\begin{equation}
I_{\mathrm{static}} = I(A^\star; \mathcal E_{\mathrm{static}} \mid Q) = H(A^\star \mid Q) - H(A^\star \mid Q, \mathcal E_{\mathrm{static}}),
\end{equation}
where $Q$ is the query and $H(\cdot)$ denotes entropy.
In contrast, an adaptive multi-turn strategy $\pi$ divides the budget across $T$ rounds as $B = \sum_{t=1}^T B_t$. At each round $t$, the agent uses prior evidence $\mathcal H_{t-1} = \{\mathcal E_1, \dots, \mathcal E_{t-1}\}$ to update its internal belief $h_{t-1}$ and selects new evidence $\mathcal E_t$ of size $B_t$ by actively exploring the graph based on current uncertainty. The updated belief $h_t$ is obtained via Bayesian inference, and the entire process forms a dynamic system:
\begin{equation}
h_t = f(h_{t-1}, \mathcal E_t, R_G).
\end{equation}
To evaluate retrieval progress, we define a Lyapunov-style potential function $V_t = H(A^\star \mid Q, \mathcal H_t)$, which quantifies the remaining uncertainty after round $t$. Each retrieval step reduces entropy by:
\begin{equation}
V_{t-1} - V_t = I(A^\star; \mathcal E_t \mid Q, \mathcal H_{t-1}),
\end{equation}
which is precisely the mutual information from $\mathcal E_t$ given past evidence.
Let $\rho_t$ denote information gain per token at round $t$:
\begin{equation}
\rho_t = \frac{I(A^\star; \mathcal E_t \mid Q, \mathcal H_{t-1})}{B_t},
\end{equation}
and define the average information density of the static strategy as:
\begin{equation}
\rho_{\mathrm{static}} = \frac{I_{\mathrm{static}}}{B}.
\end{equation}
Since the adaptive agent tailors each $\mathcal E_t$ to the current belief state and selects high-impact graph regions, it is expected that $\rho_t \ge \rho_{\mathrm{static}}$ each round. When the graph allows pruning irrelevant branches via early evidence, this inequality becomes strict with non-zero probability.
Summing over all rounds, the total information gain of the adaptive strategy satisfies:
\begin{equation}
\mathbb{E}_\pi \left[\sum_{t=1}^T I(A^\star; \mathcal E_t \mid Q, \mathcal H_{t-1})\right]
= \mathbb{E}_\pi \left[ I(A^\star; \mathcal H_T \mid Q) \right]
\ge I_{\mathrm{static}},
\end{equation}
with strict inequality under the above conditions.
From a Bayesian viewpoint, retrieval efficiency can be seen as how much uncertainty is reduced per token. Because the adaptive policy achieves a greater entropy reduction under the same budget, or requires fewer tokens to reach the same posterior certainty, it is strictly more efficient.
Moreover, by Fano’s inequality,
\begin{equation}
P_e \le \frac{H(A^\star \mid Q) - I(A^\star; \mathcal H_T \mid Q) + 1}{\log |\mathcal A|},
\end{equation}
we conclude that the lower the conditional entropy, the lower the expected error. Therefore, greater mutual information directly translates into improved answer accuracy.

In conclusion, multi-turn interaction enables the agent to reason over what has already been retrieved, selectively expanding into the most informative parts of the graph, leading to more efficient and accurate question answering.
\end{proof}
\subsection{Proof of Proposition 3}
\label{proof3}
\textbf{Proposition 3.}
\textit{End-to-end RL bridges the gap between graph-based knowledge and language.}\vspace{-2mm}
\begin{proof}
Let $G_H$ denote the graph-structured knowledge base, and $q$ a given query. The agent (parameterized by $\theta$) interacts over multiple steps, forming a trajectory $\tau = (s_1, a_1, \dots, s_T, a_T)$ where each $a_t$ is either a graph query or a natural-language output. The policy induces an answer distribution:
\begin{equation}
P_\theta(y \mid q, G_H) = \sum_{\tau:\,\text{answer}(\tau)=y} \pi_\theta(\tau \mid q, G_H).
\end{equation}
To align graph usage with answer generation, we define a trajectory-level reward:
\begin{equation}
R(\tau) = r_{\text{fmt}}(\tau) + \mathbb{I}\{r_{\text{fmt}}(\tau)=1\} \cdot r_{\text{ans}}(y_T, y_q^\star) - 1,
\end{equation}
where $r_{\text{fmt}}$ ensures proper structure (e.g., retrieve before answer), and $r_{\text{ans}}$ measures answer quality. Only valid, grounded answers receive positive reward.
The expected reward is maximized via policy gradient:
\begin{equation}
\nabla_\theta J(\theta) \propto \mathbb{E}_{\tau \sim \pi_\theta} \left[ \sum_{t=1}^T \nabla_\theta \log \pi_\theta(a_t \mid s_t; G_H) \cdot \hat{A}(\tau) \right],
\end{equation}
where $\hat{A}(\tau)$ derives from $R(\tau)$. Trajectories that retrieve the right subgraph and generate correct answers are reinforced, linking graph retrieval to accuracy.
As training progresses, the expected log-likelihood of the gold answer increases:
\begin{equation}
\mathcal{L}_\theta = - \log \sum_\tau \pi_\theta(\tau \mid q, G_H) P(y_q^\star \mid \tau),
\end{equation}
which lower-bounds the ideal $\log P^\star(y_q^\star \mid q, G_H)$. In the limit, we approach:
\begin{equation}
P_\theta(\cdot \mid q, G_H) \to P^\star(\cdot \mid q, G_H).
\end{equation}
This also manifests as a reduction in conditional entropy:
\begin{equation}
H_\theta(Y \mid Q, G_H) < H(Y \mid Q),
\end{equation}
since ungrounded answers are discouraged and graph-consistent ones are promoted. By Fano’s inequality, lower entropy implies lower error.

Thus, end-to-end RL not only learns to query the graph but also binds retrieved knowledge to answer generation, effectively bridging the gap between structure and language.
\end{proof}

\section{Graph-R1 Algorithm Details}
To illustrate the mechanism of Graph-R1, we present its full workflow in Algorithm~\ref{alg:graph-r1}, comprising three key phases:  
\textbf{(1) Hypergraph Construction.} An LLM-based extractor $\pi_{\text{ext}}$ extracts $n$-ary relational facts from the corpus $K$ to build a semantic hypergraph $\mathcal{G}_H = (V, E_H, \phi)$, where all elements are encoded via $\mathrm{Enc}(\cdot)$.  
\textbf{(2) Multi-turn Agentic Reasoning.} Given query $q$, the agent performs reflection, intent selection, and action generation over $T$ steps under policy $\pi_\theta$. Queries trigger dual-path hypergraph retrieval; answers terminate reasoning.  
\textbf{(3) End-to-end RL Optimization.} The policy is optimized using GRPO, guided by format and answer rewards. The objective $\mathcal{J}_{\text{GRPO}}$ is computed from sampled trajectories $\{\tau_i\}$ with clipped advantage-weighted updates.

\begin{algorithm}[h!t]
\small
\caption{Graph-R1: Agentic GraphRAG via End-to-end RL}
\label{alg:graph-r1}
\begin{algorithmic}[1]
\Require Query $q$, knowledge corpus $K = \{d_1, \dots, d_N\}$, policy $\pi_\theta$, reward function $R(\tau)$
\Ensure Final answer $y_q$

\vspace{1mm}
\State \textbf{// 1: Knowledge Hypergraph Construction}
\State Initialize hypergraph $\mathcal{G}_H = (V, E_H, \phi)$
\For{each document $d \in K$}
    \State Extract relational facts: $\{(h_i, \mathcal{V}_{h_i})\} \sim \pi_{\text{ext}}(d)$
    \For{each $(h_i, \mathcal{V}_{h_i})$}
        \State $E_H \gets E_H \cup \{h_i\}, \quad V \gets V \cup \mathcal{V}_{h_i}$
        \State $\phi(h_i) \gets \mathrm{Enc}(h_i), \quad \phi(v) \gets \mathrm{Enc}(v)$ for $v \in \mathcal{V}_{h_i}$
    \EndFor
\EndFor

\vspace{1mm}
\State \textbf{// 2: Multi-turn Graph Reasoning}
\State Initialize state $s_1 \gets q$, trajectory $\tau \gets \emptyset$
\For{$t = 1$ to $T$}
    \State Generate reasoning plan: $\mathbf{a}_t^{\text{think}} \sim \pi_\theta(\cdot \mid s_t)$
    \State Choose intent: $\alpha_t \sim \pi_\theta(\cdot \mid \mathbf{a}_t^{\text{think}}, s_t)$

    \If{$\alpha_t = \texttt{(answer)}$}
        \State Output answer: $\mathbf{a}_t^{\text{ans}} \sim \pi_\theta(\cdot \mid \mathbf{a}_t^{\text{think}}, s_t)$
        \State $\tau \gets \tau \cup \{(s_t, \mathbf{a}_t^{\text{query}}, \mathbf{a}_t^{\text{ans}})\}$;\quad \Return $y_q = \mathbf{a}_t^{\text{ans}}$

    \ElsIf{$\alpha_t = \texttt{(query, retrieve)}$}
        \State Generate query: $\mathbf{a}_t^{\text{query}} \sim \pi_\theta(\cdot \mid \mathbf{a}_t^{\text{think}}, s_t)$

        \State Entity retrieval:
        $\mathcal{R}_V = \operatorname*{argmax}^{k_V}_{v \in V} \text{sim}(\phi(v), \phi(V_{\mathbf{a}_t^{\text{query}}}))$
        \State Hyperedge retrieval:
        $\mathcal{R}_H = \operatorname*{argmax}^{k_H}_{h \in E_H} \text{sim}(\phi(h), \phi(\mathbf{a}_t^{\text{query}}))$
        \State Rank fusion:
        $\mathbf{a}_t^{\text{ret}} = \text{Top-}k\left(\mathcal{F}_V^* \cup \mathcal{F}_H^*,\; \text{Score}(f) = \tfrac{1}{r_V(f)} + \tfrac{1}{r_H(f)}\right)$

        \State Update state $s_{t+1} \gets s_t \cup \{(s_t, \mathbf{a}_t^{\text{think}}, \mathbf{a}_t^{\text{query}}, \mathbf{a}_t^{\text{ret}})\}$
        \State $\tau \gets \tau \cup \{(s_t, \mathbf{a}_t^{\text{think}}, \mathbf{a}_t^{\text{query}}, \mathbf{a}_t^{\text{ret}})\}$
    \EndIf
\EndFor

\vspace{1mm}
\State \textbf{// 3: End-to-end Policy Optimization (GRPO)}
\State Sample $N$ trajectories $\{\tau_i\} \sim \pi_{\theta_{\text{old}}}$
\For{each $\tau_i$}
    \State Compute reward:
    $R(\tau_i) = -1 + R_{\text{format}}(\tau_i) + \mathbb{I}\{R_{\text{format}}=1\} \cdot R_{\text{answer}}(y_T, y_q^\star)$
    \State Compute advantage:
    $\hat{A}(\tau_i) = \frac{R(\tau_i) - \mathrm{mean}(\{R(\tau_j)\})}{\mathrm{std}(\{R(\tau_j)\})}$
\EndFor

\State Update policy via GRPO:
$\mathcal{J}_{\text{GRPO}} \sim \sum_{i=1}^N \sum_{t=1}^{|\tau_i|} 
\min\left(
\rho_\theta(a_t^{(i)}) \hat{A}(\tau_i), 
\text{clip}(\rho_\theta(a_t^{(i)}), 1 \pm \epsilon) \hat{A}(\tau_i)
\right)$
\State where
$\rho_\theta(a_t^{(i)}) = \frac{\pi_\theta(a_t^{(i)} \mid s_{t-1}^{(i)})}{\pi_{\theta_{\text{old}}}(a_t^{(i)} \mid s_{t-1}^{(i)})}$

\end{algorithmic}
\end{algorithm}
\vspace{-0.5mm}

\textit{Complexity Analysis. }
Graph-R1 involves three computational components corresponding to phases. 
First, hypergraph construction scales with the total token count $T_K$ of the knowledge corpus and the number of extracted relational facts $F$, yielding complexity $O(T_K) + O(F)$. 
Second, during multi-turn reasoning, the agent performs $T$ steps of action sampling and dual-path retrieval. At each step, similarity computations over $|V|$ nodes and $|E_H|$ hyperedges with embedding dimension $d$ yield $O((|V| + |E_H|)d)$ per step. 
Third, for policy optimization, GRPO processes $N$ sampled trajectories of max length $T$, with gradient updates costing $O(NTd)$. 
Each component is computationally tractable and benefits from parallelization and localized retrieval over compact hypergraph subsets.

\section{Dataset Details}
\label{Dataset}
We conduct experiments on six widely-used RAG benchmarks selected from the FlashRAG toolkit~\citep{FlashRAG}, covering both single-hop and multi-hop question answering tasks:
\vspace{-1mm}
\begin{itemize}[leftmargin=*]
  \item \textbf{2WikiMultiHopQA (2Wiki.)} \citep{2WikiMultiHopQA}: A multi-hop dataset requiring reasoning across two Wikipedia documents.
  \item \textbf{HotpotQA} \citep{HotpotQA}: A challenging multi-hop QA dataset with sentence-level supporting facts and diverse question types.
  \item \textbf{Musique} \citep{Musique}: Multi-hop questions needing chains of inference, often involving three or more reasoning steps.
  \item \textbf{Natural Questions (NQ)} \citep{NQ}: A large-scale single-hop QA dataset grounded in real Google search questions with Wikipedia passages.
  \item \textbf{PopQA} \citep{PopQA}: An open-domain QA dataset focused on popular culture questions sourced from Wikipedia.
  \item \textbf{TriviaQA} \citep{TriviaQA}: A large-scale dataset containing trivia-style questions with distantly supervised evidence documents.
\end{itemize}
\vspace{-1mm}
To ensure consistency across datasets and maintain manageable training and evaluation workloads, we uniformly sample 5,120 instances per dataset for training and 128 instances for testing.

\section{Baseline Details}
\label{Baseline}
Our experiments compare Graph‑R1 with two groups of baselines using different backbone LLMs:
\vspace{-1.5mm}
\subsection{Baselines with \texttt{GPT‑4o‑mini}}
\begin{itemize}[leftmargin=*]
  \item \textbf{NaiveGeneration (GPT‑4o‑mini)}: Zero‑shot generation using GPT‑4o‑mini without retrieval, evaluating base model capacity.
  \item \textbf{StandardRAG (GPT‑4o‑mini)} \citep{RAG}: Chunk‑based RAG using GPT‑4o‑mini as the generator with retrieval over text chunks.
  \item \textbf{GraphRAG} \citep{GraphRAG}: Graph‑structured retrieval baseline that constructs entity graphs and performs one‑shot retrieval with GPT‑4o‑mini for answer generation.
  \item \textbf{LightRAG} \citep{LightRAG}: A lightweight GraphRAG variant that builds compact graphs for more efficient retrieval and GPT‑4o‑mini generation.
  \item \textbf{PathRAG} \citep{PathRAG}: Retrieval via path-based pruning on entity graphs, followed by GPT‑4o‑mini answer synthesis.
  \item \textbf{HippoRAG2} \citep{HippoRAG2}: Hierarchical path planner over knowledge graphs to improve retrieval efficiency, with GPT‑4o‑mini used for generation.
  \item \textbf{HyperGraphRAG} \citep{HyperGraphRAG}: Constructs n‑ary relational hypergraphs to support a single retrieval step, and uses GPT‑4o‑mini for answer writing.
\end{itemize}
\vspace{-1.5mm}
\subsection{Baselines with \texttt{Qwen2.5} (1.5B, 3B, 7B)}
\begin{itemize}[leftmargin=*]
  \item \textbf{NaiveGeneration}: Direct generation by Qwen2.5 given the question prompt, without any retrieval, serving as a lower bound baseline.
  \item \textbf{StandardRAG} \citep{RAG}: Classic chunk-based retrieval-augmented generation pipeline with semantic retriever and Qwen2.5 decoder.
  \item \textbf{SFT} \citep{SFT}: Supervised fine-tuning of Qwen2.5 on QA pairs, without multi-turn reasoning or reinforcement optimization.
  \item \textbf{R1} \citep{GRPO}: A GRPO-trained policy that generates final answers directly from question prompts without retrieval, optimized only on answer quality.
  \item \textbf{Search‑R1} \citep{Search-R1}: A multi-turn chunk-based retrieval method trained with GRPO, capable of iterative query refinement and retrieval under a unified policy.
  \item \textbf{R1‑Searcher} \citep{R1-Searcher}: A two-stage GRPO-based method with chunk-based retrieval: first using only format-level rewards to produce structured traces, then adding answer rewards.
\end{itemize}

\section{Evaluation Details}
\label{Evaluation}
\begin{figure}[h]
\vspace{-0.5mm}
\centering
\includegraphics[width=\linewidth]{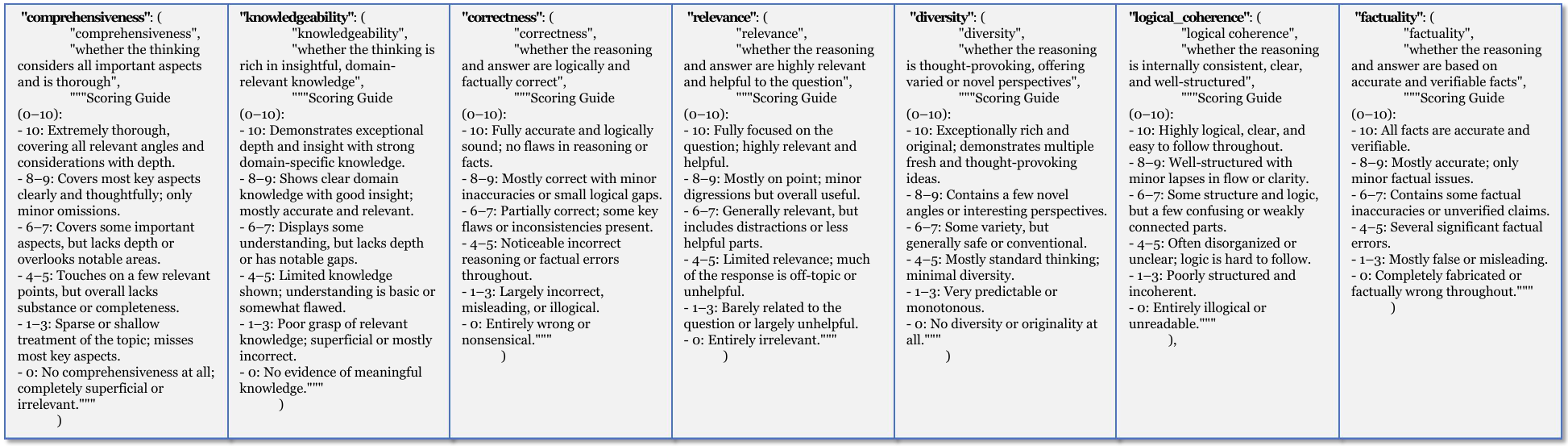}
\vspace{-6mm}
\caption{\label{prompt3}
Seven Dimensions for Generation Evaluation.}
\vspace{-2mm}
\end{figure}

\begingroup
\setlength{\abovedisplayskip}{2pt}
\setlength{\belowdisplayskip}{2pt}
\setlength{\abovedisplayshortskip}{2pt}
\setlength{\belowdisplayshortskip}{2pt}

Inspired by~\citet{HyperGraphRAG,Search-R1}, we evaluate model performance using four metrics:

\noindent
\begin{minipage}[t]{0.48\textwidth}
\textbf{(i) Exact Match (EM). }
EM measures whether the predicted answer exactly matches the ground truth. Let $\text{norm}(\cdot)$ denote the normalization function:
\begin{equation}
\small
\text{EM} = \frac{1}{N} \sum_{i=1}^{N} \mathbb{I} \left\{ \text{norm}(y_i) = \text{norm}(y_i^\star) \right\}.
\end{equation}
\textbf{(ii) F1 Score. } 
The F1 score measures the token-level overlap between the predicted answer $y_i$ and the ground-truth answer $y^\star$ using the harmonic mean of precision and recall:
\begin{equation}
\small
\text{F1} = \frac{1}{N} \sum_{i=1}^{N} \frac{2 \cdot |\text{tokens}(y_i) \cap \text{tokens}(y_i^\star)|}{|\text{tokens}(y_i)| + |\text{tokens}(y_i^\star)|}.
\end{equation}
\end{minipage}
\hfill
\begin{minipage}[t]{0.48\textwidth}
\textbf{(iii) Retrieval Similarity (R-S). }
R-S assesses semantic similarity between retrieved $k_{\text{retr}}^{(i)}$ and gold knowledge $k_{\text{retr}}^{(i)}$. Let $\text{Enc}(\cdot)$ be the semantic embedding function:
\begin{equation}
\small
\text{R-S} = \frac{1}{N} \sum_{i=1}^{N} \cos\left(\text{Enc}(k_{\text{retr}}^{(i)}), \text{Enc}(k_{\text{gold}}^{(i)})\right).
\end{equation}
\textbf{(iv) Generation Evaluation (G-E). }
G-E reflects generation quality. Let $s_{i,d}$ be GPT-4o-mini scores across 7 criteria (Figure~\ref{prompt3}):
\begin{equation}
\small
\text{G-E} = \frac{1}{N} \sum_{i=1}^{N} \left( \frac{1}{7} \sum_{d=1}^{7} s_{i,d}\right).
\end{equation}
\end{minipage}
\endgroup

\section{Implementation Details}
\label{Implementation}
As shown in Table~\ref{tab:hyperparams}, we summarize the detailed hyperparameter configurations used throughout our experiments, including model backbone, input limits, training configuration, and retrieval setup.
\begin{table}[h]
\centering
\fontsize{7pt}{7.5pt}\selectfont
\caption{\label{tab:hyperparams}
Hyperparameter settings for baselines and Graph-R1.}
\setlength{\tabcolsep}{2.8mm}{
\begin{tabular}{lcccccc}
\toprule
\textbf{Method} & \textbf{Backbone} & \textbf{Batch Size} & \textbf{Max Length} & \textbf{Top-K} & \textbf{Algo} & \textbf{Epochs} \\
\midrule
NaiveGeneration     & Qwen2.5 / GPT-4o-mini & --   & $\infty$ & N/A  & --     & --  \\
StandardRAG         & Qwen2.5 / GPT-4o-mini & --   & $\infty$ & 5 Chunks  & --     & --  \\
GraphRAG            & GPT-4o-mini           & --   & $\infty$ & 60 & --     & --  \\
LightRAG            & GPT-4o-mini           & --   & $\infty$ & 60  & --     & --  \\
PathRAG             & GPT-4o-mini           & --   & $\infty$ & 60  & --     & --  \\
HippoRAG2           & GPT-4o-mini           & --   & $\infty$ & 60  & --     & --  \\
HyperGraphRAG       & GPT-4o-mini           & --   & $\infty$ & 60  &  --     & --  \\
SFT                 & Qwen2.5 (1.5B, 3B, 7B)              & 16   & 4096 & N/A  & LoRA     & 3   \\
\rowcolor{R1!15} R1                  & Qwen2.5 (1.5B, 3B, 7B)               & 128   & 4096 & N/A  & GRPO   & 1   \\
\rowcolor{Search-R1!15} Search-R1           & Qwen2.5 (1.5B, 3B, 7B)               & 128   & 4096 & 5 Chunks / Turn & GRPO & 1 \\
\rowcolor{R1-Searcher!15} R1-Searcher         & Qwen2.5 (1.5B, 3B, 7B)               & 128   & 4096 & 5 Chunks / Turn & GRPO & 1 \\
\rowcolor{Graph-R1!15}
Graph-R1 (ours)     & Qwen2.5 (1.5B, 3B, 7B)               & 128   & 4096 & 5 / Turn & GRPO & 1 \\
\bottomrule
\end{tabular}}
\vspace{-1mm}
% \caption{\label{tab:hyperparams}
% Hyperparameter settings for baselines and Graph-R1.}
\vspace{-2mm}
\end{table}

\section{Case Study Analysis}

\begin{table}[h]
\centering
\caption{\label{T4}
Case study on generation quality under a query, comparing NaiveGeneration, StandardRAG, HyperGraphRAG based on GPT-4o-mini, with R1 (7B), Search-R1 (7B), and Graph-R1 (7B).}
\includegraphics[width=0.96\linewidth]{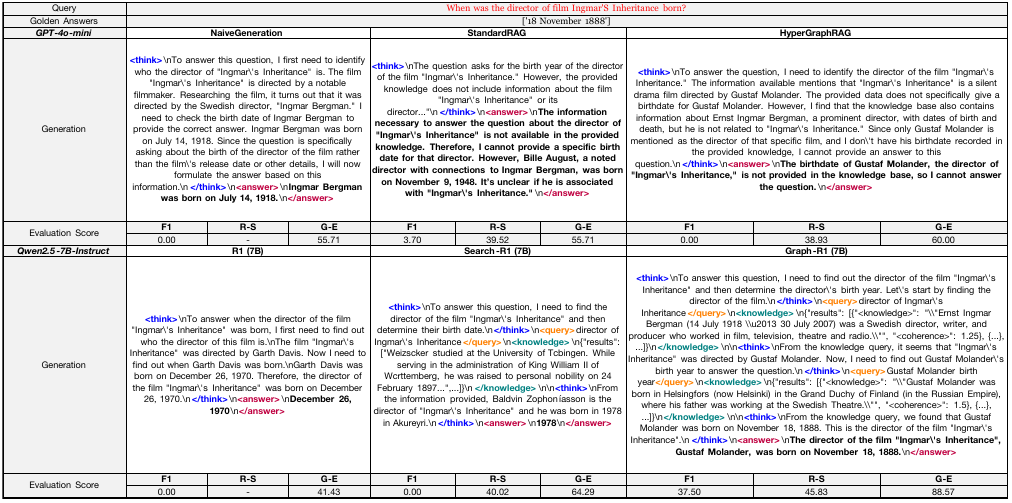}
% \vspace{-1.5mm}
% \caption{\label{T4}
% Case study on generation quality under a query, comparing NaiveGeneration, StandardRAG, HyperGraphRAG based on GPT-4o-mini, with R1 (7B), Search-R1 (7B), and Graph-R1 (7B).}
% \vspace{-1.5mm}
\end{table}

As shown in Table~\ref{T4}, NaiveGeneration and R1 fail to provide the correct answer, and both StandardRAG and HyperGraphRAG also fall short despite using structured prompts. Search-R1, though RL-enhanced, shows limited improvement due to weak retrieval grounding. In contrast, Graph-R1 accurately identifies both the director and birthdate, achieves the highest G-E score (88.57), and demonstrates that RL is more effective with graph-based knowledge interaction.

\section{Analysis of Graph-R1’s Generalizability on O.O.D. Settings}
\begin{figure*}[h]
\footnotesize
\centering
\begin{subfigure}[t]{0.24\textwidth}
  \centering
  \includegraphics[width=\linewidth]{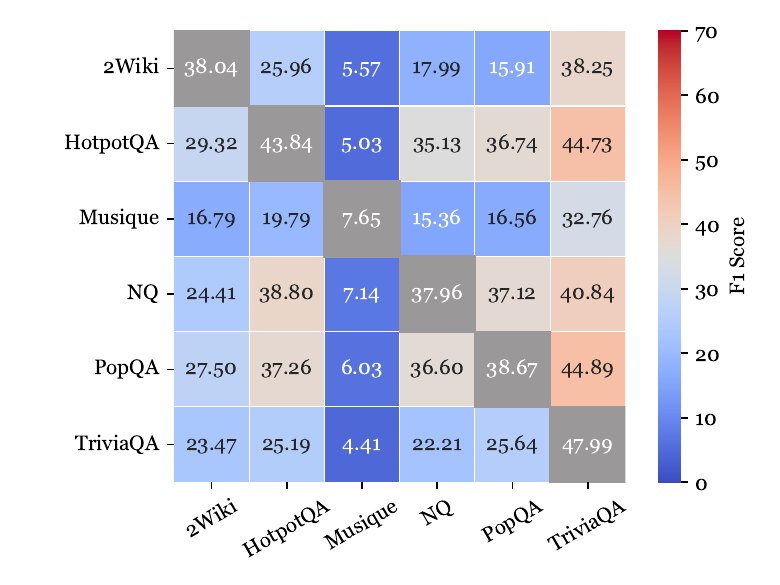}
  \vspace{-5mm}
  \caption{Datasets}
  \label{F8a}
\end{subfigure}\hspace{-3mm}
\begin{subfigure}[t]{0.24\textwidth}
  \centering
  \includegraphics[width=\linewidth]{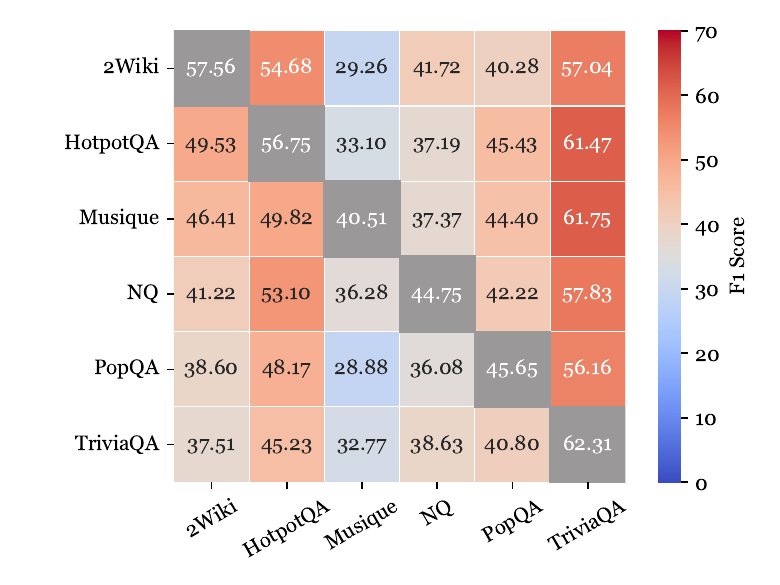}
  \vspace{-5mm}
  \caption{Parameters}
  \label{F8b}
\end{subfigure}\hspace{-1mm}
\begin{subfigure}[t]{0.24\textwidth}
  \centering
  \includegraphics[width=\linewidth]{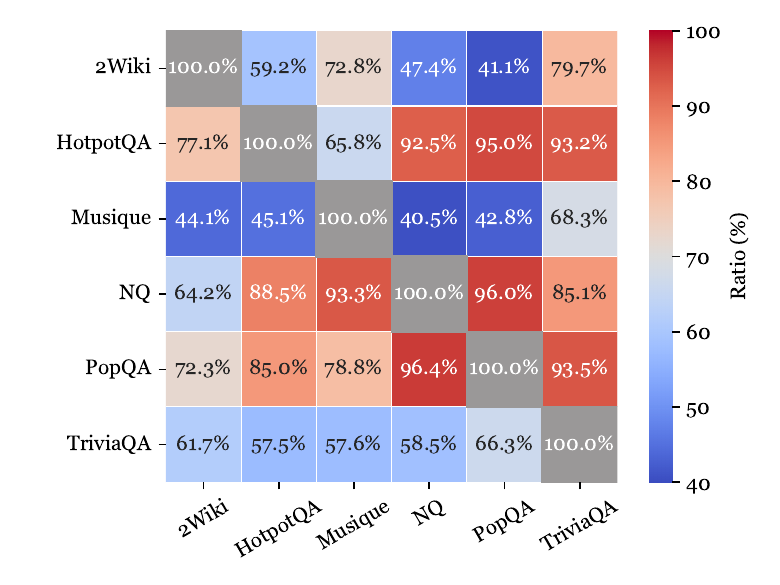}
  \vspace{-5mm}
  \caption{Qwen3}
  \label{F8c}
\end{subfigure}\hspace{-3mm}
\begin{subfigure}[t]{0.24\textwidth}
  \centering
  \includegraphics[width=\linewidth]{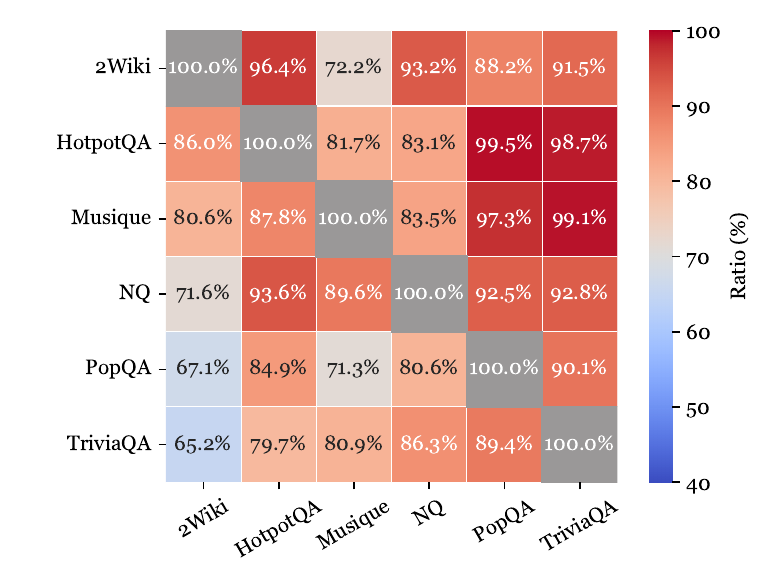}
  \vspace{-5mm}
  \caption{Algorithms}
  \label{F8d}
\end{subfigure}

\vspace{-1mm}
\caption{F1 comparison and performance ratios across six datasets under O.O.D. cross-validation.}
\label{F8}
\end{figure*}

As shown in Figure~\ref{F8}, to verify generalization, we conduct O.O.D. cross-validation for Search-R1 (3B) \& Graph-R1 (3B) across six datasets: (a-b) F1 comparison, and (c-d) O.O.D.-to-I.I.D. ratios.

\textbf{F1 Performance Across Datasets. }
Figures~\ref{F8a} and~\ref{F8b} show that Graph-R1 outperforms Search-R1 on six datasets in O.O.D. validation, with notable gains on NQ and TriviaQA. Its multi-turn interaction with hypergraph retrieval ensures more stable performance under distribution shifts.

\textbf{Robust Generalization Ability. }
Figures~\ref{F8c} and~\ref{F8d} show that Graph-R1 achieves higher O.O.D.-to-I.I.D. ratios than Search-R1, often above 85\% and exceeding 90\% in some cases, reflecting its strong robustness and cross-domain generalizability via end-to-end RL over knowledge hypergraph.

\clearpage
\section{Details of Knowledge HyperGraph Construction}
\begin{figure}[h]
\centering
\includegraphics[width=16.5cm]{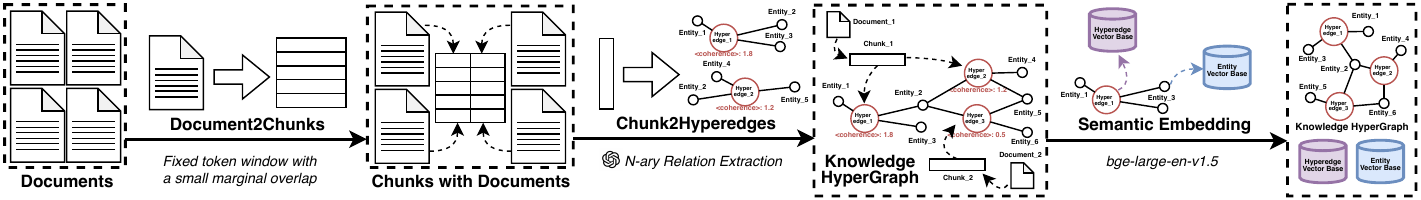}
\caption{Knowledge HyperGraph Construction in Graph-R1.}
\label{F11}
\end{figure}
As shown in Figure~\ref{F11}, Graph-R1 constructs the Knowledge HyperGraph through a streamlined yet expressive three-stage pipeline.

\textbf{First}, we segment raw documents into fixed-size textual units to maintain stable local semantics. Specifically, each document is divided using a 1200-token window with a 50-token overlap, ensuring that adjacent chunks retain sufficient contextual continuity. This windowing strategy preserves semantically meaningful boundaries while preventing the loss of cross-sentence information.

\textbf{Second}, each chunk is processed using a GPT-4o-mini–based n-ary relation extraction module. We adopt the carefully designed prompt shown in Appendix~\ref{AppendixA1}, Figure~\ref{prompt1}, which guides the model to identify sets of entities and the higher-order relations binding them. Each extracted relational fact is represented as a hyperedge, and we assign a coherence score that reflects the internal semantic consistency and reliability of the extracted n-ary relation. This step transforms linear text into a structured hypergraph where nodes represent entities and hyperedges capture rich n-ary facts.

\textbf{Third}, to enable semantic retrieval and reasoning over the constructed hypergraph, we generate dense embeddings for both entities and hyperedges using the large-scale embedding model bge-large-en-v1.5. The resulting entity vector base and hyperedge vector base form the foundation of Graph-R1’s retrieval module, allowing subsequent reasoning steps.

Compared with the original HyperGraphRAG implementation, we introduce several architectural and engineering optimizations, particularly in parallelization, prompt efficiency, storage layout, and vectorization. These improvements reduce token consumption and computational overhead, yielding a more efficient hypergraph construction pipeline, as evidenced in Table~\ref{T3} of Section~\ref{5.4}.

\section{More Experimental Results on O.O.D. Settings}

To rigorously assess the out-of-domain (O.O.D.) robustness and cross-domain generalization of Graph-R1, we evaluate the model under three distinct training regimes, each designed to isolate a different aspect of domain transferability. These settings provide a comprehensive picture of how retrieval structures (chunks vs. hypergraphs) behave when training and testing domains differ.

\textbf{(i) In-Domain (I.I.D.) Training Setting. }
The first setting follows the same configuration as in Table~\ref{T2}.
Here, each model is trained exclusively on the training split of its own dataset, meaning the supervision is fully aligned with the target domain. This represents the upper-bound scenario in which rich in-domain training data are available, allowing us to evaluate whether Graph-R1 can surpass Search-R1 even without requiring cross-domain generalization.

\textbf{(ii) Single O.O.D. Training Setting. }
The second setting corresponds to the cross-dataset training paradigm illustrated in Figures~\ref{F8}.
For each target dataset, the model is trained not on its own data, but on the training split of a different dataset. We use the results and compute the average performance. This setting stresses the ability to extract transferable information and tests whether hypergraph-based retrieval yields more robust inductive biases than chunk-based retrieval under dataset shift.

\textbf{(iii) Combined-Dataset Training Setting. }
The third regime evaluates the model in a mixed-domain learning scenario.
We first merge all six datasets into a single training pool and then uniformly downsample 1/6 of the combined data to ensure that the total training data volume remains identical to the I.I.D. setting. The resulting training set contains balanced signals drawn from diverse domains while strictly controlling for data quantity. This design allows us to examine whether Graph-R1 can exploit heterogeneous multi-domain relational structures to improve generalization, while ruling out gains attributable merely to increased training size.

\subsection{Results and Analysis on Six Open-Domin Datasets}

\begin{table*}[h]
\centering
\fontsize{7pt}{7.5pt}\selectfont
\caption{\label{T5}
Comparison between Search-R1 (3B) and Graph-R1 (3B) across six multi-hop and open-domain datasets under three training regimes. Best results are in \textbf{bold} and second in \underline{underline}.}
\setlength{\tabcolsep}{1.5mm}{
\begin{tabular}{l cc cc cc cc cc cc cc}
\toprule
\multirow{2}{*}{\textbf{Method}} 
& \multicolumn{2}{c}{\textbf{2Wiki.}} 
& \multicolumn{2}{c}{\textbf{HotpotQA}} 
& \multicolumn{2}{c}{\textbf{Musique}} 
& \multicolumn{2}{c}{\textbf{NQ}} 
& \multicolumn{2}{c}{\textbf{PopQA}} 
& \multicolumn{2}{c}{\textbf{TriviaQA}}
& \multicolumn{2}{c}{\textbf{Avg.}} 
\\
\cmidrule(lr){2-3}
\cmidrule(lr){4-5}
\cmidrule(lr){6-7}
\cmidrule(lr){8-9}
\cmidrule(lr){10-11}
\cmidrule(lr){12-13}
\cmidrule(lr){14-15}
& \textbf{EM} & \textbf{F1} 
& \textbf{EM} & \textbf{F1} 
& \textbf{EM} & \textbf{F1} 
& \textbf{EM} & \textbf{F1} 
& \textbf{EM} & \textbf{F1} 
& \textbf{EM} & \textbf{F1}
& \textbf{EM} & \textbf{F1} \\
\midrule
\multicolumn{15}{c}{\textit{\textbf{Qwen2.5-3B-Instruct} (Results When Trained on Its Own I.I.D. Dataset)}} \\
\rowcolor{Search-R1!15} Search-R1 (3B) & 31.25 & 38.04 & 38.28 & 43.84 & 3.91 & 7.65 & 24.22 & 37.96 & 33.59 & 38.67 & 40.62 & 47.99 & 28.65 & 35.69 \\
\rowcolor{Graph-R1!15} Graph-R1 (3B) & \textbf{50.00} & \textbf{57.56} & \textbf{50.78} & \textbf{56.75} & \textbf{32.81} & \textbf{40.51} & \textbf{30.47} & \textbf{44.75} & \underline{37.50} & \textbf{45.65} & \textbf{53.13} & \textbf{62.31} & \textbf{42.45} & \textbf{51.26} \\
\midrule
\multicolumn{15}{c}{\textit{\textbf{Qwen2.5-3B-Instruct} (Average Performance When Trained on a Single O.O.D. Dataset)}} \\
\rowcolor{Search-R1!15} Search-R1 (3B) & -- & 24.30 & -- & 29.40 & -- & 5.64 & -- & 25.46 & -- & 26.59 & -- & 40.29 & -- & 25.28 \\
\rowcolor{Graph-R1!15} Graph-R1 (3B) & -- & 42.65 & -- & 50.20 & -- & 32.06 & -- & 38.20 & -- & 42.63 & -- & \underline{58.85} & -- & 44.10 \\
\midrule
\multicolumn{15}{c}{\textit{\textbf{Qwen2.5-3B-Instruct} (Results When Trained on Combined Datasets)}} \\
\rowcolor{Search-R1!15} Search-R1 (3B) & 25.00 & 29.68 & 36.72 & 41.59 & 10.16 & 15.48 & 21.09 & 34.78 & 35.16 & 39.07 & 42.97 & 50.92 & 28.52 & 35.25 \\
\rowcolor{Graph-R1!15} Graph-R1 (3B) & \underline{42.97} & \underline{51.07} & \underline{46.88} & \underline{54.97} & \underline{32.03} & \underline{38.29} & \underline{27.34} & \underline{41.00} & \textbf{38.28} & \underline{45.32} & \underline{50.00} & 58.49 & \underline{39.58} & \underline{48.19} \\
\bottomrule
\end{tabular}
}
% \caption{\label{T5}
% Comparison between Search-R1 (3B) and Graph-R1 (3B) across six multi-hop and open-domain datasets under three training regimes. Best results are in \textbf{bold} and second in \underline{underline}.}
\end{table*}

Table~\ref{T5} reports the performance of Search-R1 and Graph-R1 under three training regimes (I.I.D., single O.O.D., and combined), covering six multi-hop and open-domain datasets.

First, Graph-R1 achieves consistent and clear improvements over Search-R1 across all datasets and all training settings. This confirms the advantage of hypergraph-structured retrieval, which provides more coherent and relationally grounded evidence than chunk-based retrieval, leading to stronger reasoning and higher EM/F1 scores.

Second, comparing the three regimes, the results follow a stable pattern: I.I.D. training performs best, combined-dataset training ranks second, and single O.O.D. training performs worst. This indicates that dataset-aligned supervision remains the most effective, but exposure to balanced multi-dataset data is still substantially better than training on a completely mismatched dataset.

\subsection{Additional O.O.D. Results on Five Specialized Domains}

\begin{table*}[h]
\centering
\fontsize{7pt}{7.5pt}\selectfont
\caption{\label{T6}
Domain-wise results on five specialized datasets. Results of GPT-4o-mini baselines are reported from the HyperGraphRAG paper~\citep{HyperGraphRAG}. Best results are in \textbf{bold} and second in \underline{underline}. }
\setlength{\tabcolsep}{2.1mm}{
\begin{tabular}{l cc cc cc cc cc cc cc}
\toprule
\multirow{2}{*}{\textbf{Method}} 
& \multicolumn{2}{c}{\textbf{Medicine}} 
& \multicolumn{2}{c}{\textbf{Agriculture}} 
& \multicolumn{2}{c}{\textbf{CS}} 
& \multicolumn{2}{c}{\textbf{Legal}} 
& \multicolumn{2}{c}{\textbf{Mix}} 
& \multicolumn{2}{c}{\textbf{Avg.}} \\
\cmidrule(lr){2-3}
\cmidrule(lr){4-5}
\cmidrule(lr){6-7}
\cmidrule(lr){8-9}
\cmidrule(lr){10-11}
\cmidrule(lr){12-13}
& \textbf{EM} & \textbf{F1}
& \textbf{EM} & \textbf{F1}
& \textbf{EM} & \textbf{F1}
& \textbf{EM} & \textbf{F1}
& \textbf{EM} & \textbf{F1}
& \textbf{EM} & \textbf{F1} \\
\midrule
\multicolumn{13}{c}{\textbf{\textit{GPT-4o-mini}}} \\
NaiveGeneration & -- & 12.89 & -- & 12.74 & -- & 18.65 & -- & 21.64 & -- & 16.93 & -- & 16.57 \\
StandardRAG     & -- & 27.90 & -- & 27.43 & -- & 28.93 & -- & 37.34 & -- & 43.20 & -- & 32.96 \\
GraphRAG        & -- & 17.60 & -- & 21.28 & -- & 23.33 & -- & 30.11 & -- & 19.27 & -- & 22.32 \\
LightRAG        & -- & 12.79 & -- & 18.24 & -- & 22.72 & -- & 31.64 & -- & 27.03 & -- & 22.48 \\
PathRAG         & -- & 14.94 & -- & 21.30 & -- & 26.73 & -- & 31.29 & -- & 37.07 & -- & 26.27 \\
HippoRAG2       & -- & 21.34 & -- & 12.63 & -- & 17.34 & -- & 18.53 & -- & 21.53 & -- & 18.27 \\
HyperGraphRAG   & -- & \underline{35.35} & -- & \underline{33.89} & -- & \underline{31.30} & -- & \textbf{43.81} & -- & \underline{48.71} & -- & \underline{38.61} \\
\midrule
\multicolumn{13}{c}{\textit{\textbf{Qwen2.5-3B-Instruct} (Zero-Shot Results Without Domain Training)}} \\
\rowcolor{Search-R1!15} Search-R1 (3B) & \underline{12.70} & 21.84 & \underline{11.33} & 17.65 & \underline{17.97} & 25.39 & \underline{14.65} & 24.84 & \underline{23.83} & 31.90 & \underline{16.10} & 24.32 \\
\rowcolor{Graph-R1!15} Graph-R1 (3B)  & \textbf{24.22} & \textbf{37.18} & \textbf{25.98} & \textbf{37.39} & \textbf{27.54} & \textbf{37.45} & \textbf{21.48} & \underline{33.66} & \textbf{38.48} & \textbf{50.60} & \textbf{27.54} & \textbf{39.26} \\
\bottomrule
\end{tabular}
}
% \caption{\label{T6}
% Domain-wise results on five specialized datasets. Results of GPT-4o-mini baselines are reported from the HyperGraphRAG paper. Best results are in \textbf{bold} and second in \underline{underline}. }
\end{table*}

To further assess zero-shot generalization, Table~\ref{T6} evaluates the model, trained only under the combined setting, on five specialized domain datasets introduced in HyperGraphRAG. For GPT-4o-mini baselines, we report the original results from the HyperGraphRAG paper.

Across Medicine, Agriculture, CS, Legal, and the Mixed domain, Graph-R1 (3B) consistently surpasses Search-R1 by sizeable margins in both EM and F1. Notably, Graph-R1 delivers strong improvements even in knowledge-intensive fields (e.g., Medicine and CS), where structured multi-entity relationships are crucial. These findings suggest that the relational inductive biases encoded by the hypergraph representation transfer effectively across specialized verticals not seen during training.

Collectively, these experiments demonstrate that Graph-R1 offers superior cross-domain robustness, strong transferability to unseen fields, and stable zero-shot performance, significantly outperforming chunk-based RAG baselines across both horizontal (general) and vertical (specialized) domains.
\clearpage

\end{document}